\documentclass[10pt,journal,compsoc]{IEEEtran}

\usepackage{epsfig,graphicx,amsfonts,xcolor,multirow,amsmath,hyperref,booktabs,balance,ulem}
\usepackage[nocompress]{cite}
\usepackage{mycommand}

\begin{document}

\title{Constructing Stronger and Faster Baselines for Skeleton-based Action Recognition}

\author{
  Yi-Fan~Song,
  Zhang~Zhang,~\IEEEmembership{Member,~IEEE,}\\
  Caifeng~Shan,~\IEEEmembership{Senior~Member,~IEEE,}
  and~Liang~Wang,~\IEEEmembership{Fellow,~IEEE}

  \IEEEcompsocitemizethanks{
    \IEEEcompsocthanksitem Yi-Fan Song, Zhang Zhang, and Liang Wang are with the School of Artificial Intelligence, University of Chinese Academy of Sciences (UCAS), Beijing 100190, China, and also with the Center for Research on Intelligent Perception and Computing (CRIPAC), National Laboratory of Pattern Recognition (NLPR), Institute of Automation, Chinese Academy of Sciences (CASIA), Beijing 100190, China. (Email: yifan.song@cripac.ia.ac.cn, zzhang@nlpr.ia.ac.cn, wangliang@nlpr.ia.ac.cn)
    \IEEEcompsocthanksitem Caifeng Shan is with the College of Electrical Engineering and Automation, Shandong University of Science and Technology (SDUST), Qingdao 266590, China, and also with the Artificial Intelligence Research, Chinese Academy of Sciences (CAS-AIR), Beijing 100190, China. (Email: caifeng.shan@gmail.com)
  }
}

\markboth{Journal of IEEE Transactions on Pattern Analysis and Machine Intelligence}
{Song \MakeLowercase{\textit{et al.}}: Constructing Stronger and Faster Baselines for Skeleton-based Action Recognition}

\IEEEtitleabstractindextext{
  \begin{abstract}
    One essential problem in skeleton-based action recognition is how to extract discriminative features over all skeleton joints. However, the complexity of the recent State-Of-The-Art (SOTA) models for this task tends to be exceedingly sophisticated and over-parameterized. The low efficiency in model training and inference has increased the validation costs of model architectures in large-scale datasets. To address the above issue, recent advanced separable convolutional layers are embedded into an early fused Multiple Input Branches (MIB) network, constructing an efficient Graph Convolutional Network (GCN) baseline for skeleton-based action recognition. In addition, based on such the baseline, we design a compound scaling strategy to expand the model's width and depth synchronously, and eventually obtain a family of efficient GCN baselines with high accuracies and small amounts of trainable parameters, termed EfficientGCN-Bx, where ''x'' denotes the scaling coefficient. On two large-scale datasets, \ie, NTU RGB+D 60 and 120, the proposed EfficientGCN-B4 baseline outperforms other SOTA methods, \eg, achieving {\bf 92.1\%} accuracy on the cross-subject benchmark of NTU 60 dataset, while being {\bf 5.82$\times$} smaller and {\bf 5.85$\times$} faster than MS-G3D, which is one of the SOTA methods. The source code in PyTorch version and the pretrained models are available at \href{https://github.com/yfsong0709/EfficientGCNv1}{https://github.com/yfsong0709/EfficientGCNv1}.
  \end{abstract}

}

\maketitle
\IEEEpeerreviewmaketitle

\IEEEraisesectionheading{
  \section{Introduction}
  \label{sec:introduction}
}

\IEEEPARstart{H}{uman} action recognition becomes increasingly crucial and achieves promising progress in various applications during the past decade, such as video surveillance, human-computer interaction, video retrieval and so on \cite{poppe2010survey,aggarwal2011human,weinland2011survey}. One essential problem in human action recognition is how to extract discriminative and rich features to fully describe the variations of spatial configurations and temporal dynamics in human actions.

Currently, skeleton-based representations are very popular for human action recognition, as human skeletons provide a compact data form to depict dynamic changes in human body movements \cite{johansson1973visual}. Skeleton data is a time series of 3D coordinates of multiple skeleton joints, which can be either estimated from 2D images by pose estimation methods \cite{cao2017realtime} or directly collected by multimodal sensors such as Kinect \cite{zhang2012microsoft}. Moreover, compared to conventional RGB based action recognition methods, skeleton-based representations are more robust to the variations of illumination, camera viewpoints and other background changes. These merits inspire researchers to develop various methods to explore informative features from skeleton motion sequences for action recognition.

The development of skeleton-based action recognition can be divided mainly into two phases. In early years, conventional methods adopt Recurrent Neural Network (RNN)-based or Convolutional Neural Network (CNN)-based models to analyze skeleton sequences. For example, Du \etal \cite{du2015hierarchical} employ a hierarchical bidirectional RNN to capture rich dependencies between different body parts. Li \etal \cite{li2017skeleton} design a simple yet effective CNN architecture for action classification from trimmed skeleton sequences. In recent years, due to the greatly expressive power for depicting structural data, graph-based models \cite{kipf2016semi,li2018adaptive} have been proposed for modeling dynamic skeleton sequences. Yan \etal \cite{yan2018spatial} firstly propose a Spatial Temporal Graph Convolutional Networks (ST-GCN) for skeleton-based action recognition, after that increasing number of studies \cite{zhang2019graph,shi2019two,song2019richly} are reported based on GCN models.

Nevertheless, for learning discriminative and rich features from skeleton sequences, current State-Of-The-Art (SOTA) models are often exceedingly sophisticated and over-parameterized, where the network often contains a multi-stream architecture with a large number of model parameters, which leads to a complicated training procedure and high computational cost (and thus low inference speed). For example, the 2s-AGCN in \cite{shi2019two} contains about 6.94 million parameters, and it takes nearly 4 GPU-days for model training on the NTU RGB+D 60 dataset \cite{shahroudy2016ntu}. And the DGNN \cite{shi2019skeleton} contains more than 26 million parameters, which makes it very hard to do parameter tuning on large-scale datasets. The high model complexity has seriously limited the development of skeleton-based action recognition, while there are few literatures on this issue.

To tackle this problem, some efforts are made in this paper to extremely reduce the redundant trainable parameters while maintaining the model performance. Firstly, an early fused Multiple Input Branches (MIB) architecture is constructed to capture rich features from both spatial configurations and temporal dynamics of joints in skeleton sequences. The early fusion strategy has been widely used to fuse multi-modal information for various visual tasks, such as Video Question Answering (VQA) \cite{huang2020location}. In this paper, we emphasize that the proposed MIB mainly aims to reduce the model parameters and computational costs of multi-stream GCN models for more efficient skeleton-based action recognition. In details, three input branches including joint positions (relative and absolute), motion velocities (one or two temporal steps), and bone features (lengths and angles) are fused in the early stage of the whole network, rather than the conventional late fusion at score layer in most multi-stream GCN models \cite{shi2019two,liu2020disentangling,chen2021multi}. The optimal fusion stage is chosen by exhaustive search (see Sec.~\ref{sssec:compare_fusion}). To the best of our knowledge, it is the first time to investigate the impacts of different fusion locations and assess the optimal fusion layer in GCN-based skeleton action recognition.

Secondly, besides the Basic Layer (BasicLayer) proposed in ST-GCN \cite{yan2018spatial}, we extend four kinds of convolutional layers in CNN, \ie, Bottleneck Layer (BottleLayer) \cite{he2016deep}, Separable Layer (SepLayer) \cite{howard2017mobilenets}, Expanded Separable Layer (EpSepLayer) \cite{sandler2018mobilenetv2}, and Sandglass Layer (SGLayer) \cite{zhou2020rethinking}, to the GCN network for extracting temporal dynamics and compressing the model size. These four layers can obviously reduce the amount of parameter tuning costs in training, and accelerate the model inference in testing.

\begin{figure}[t]
  \centerline{\includegraphics[width=9cm]{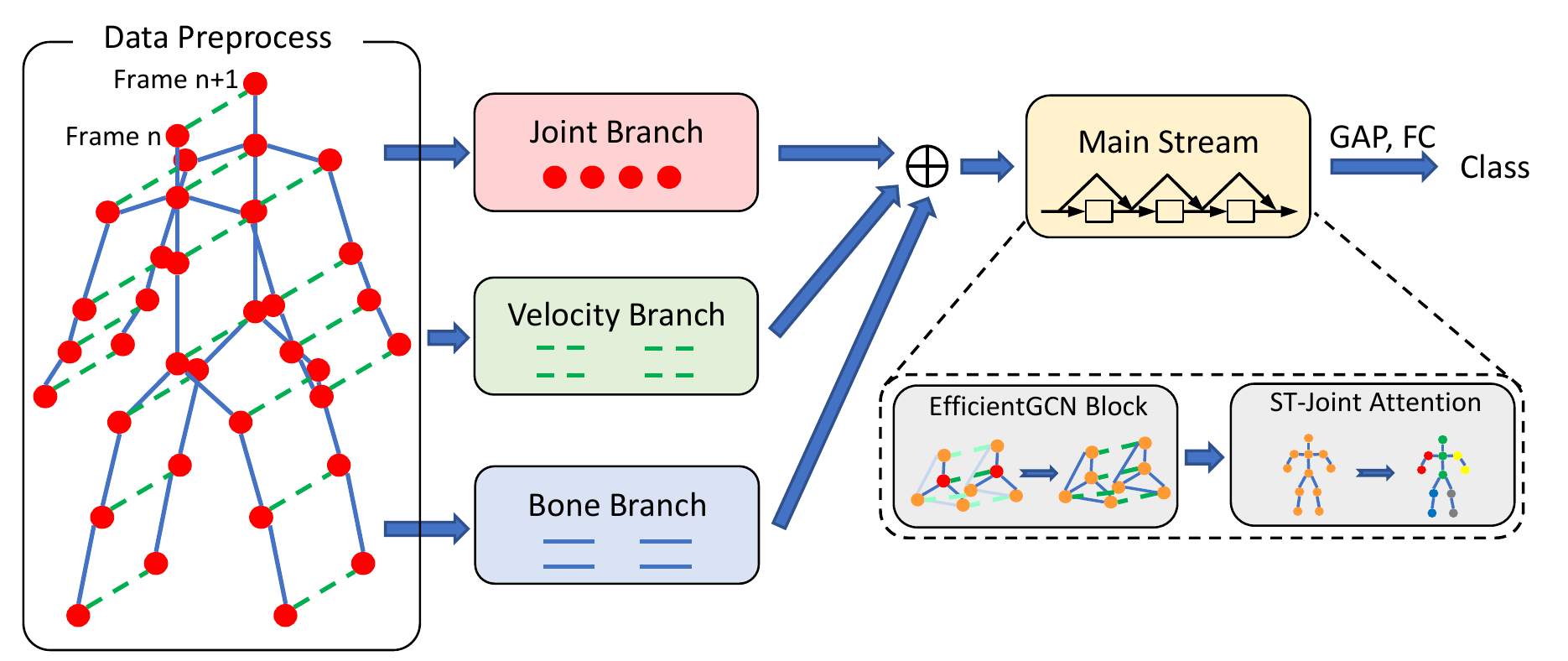}}
  \vspace{-0.4cm}
  \caption{The overall pipeline of our approach, where $\oplus$ represents concatenation operation, GAP and FC denote the Global Average Pooling operation and Fully Connected layer, respectively. \bv}\label{fig:pipeline}
  \vspace{-0.4cm}
\end{figure}

Thirdly, in order to determine the structural hyper-parameters for each block, we resort to the compound scaling method proposed by Tan and Le \cite{tan2019efficientnet}, which uniformly scales the network width, depth and resolution with a set of fixed scaling coefficients. Due to its high efficiency, the original compound scaling strategy has gradually become a popular baseline constructing method for many visual recognition tasks. However, it is quite hard to directly utilize the compound scaling strategy in GCN-based models because the resolution scaling in original EfficientNet \cite{tan2019efficientnet} cannot be implemented on graph data explicitly. To address this issue, we modify the original scaling strategy to adapt to graph data, by removing the resolution scaling factor and rebuilding the constraint between width and depth factors. Such a strategy improves the model performance in an efficient way, bringing a competitive model accuracy with significantly fewer model parameters and lower computational cost than other GCN-based SOTA methods.

Finally, for more accurate recognition, inspired by Hou \etal \cite{hou2021coordinate}, an attention module, named Spatial Temporal Joint Attention (ST-JointAtt), is proposed and inserted into each block of the model. This attention module aims to find the most essential joints from the whole skeleton sequence, and eventually enhances the model ability to extract discriminative features. Compared to other attention modules such as STC-attention (STCAtt) \cite{shi2020skeleton} and Part-wise Attention (PartAtt) \cite{song2020stronger}, this new module jointly deals with the spatial and temporal attentions, while STCAtt is asynchronous and PartAtt ignores the temporal differences. In addition, compared with the previous PartAtt module, the proposed ST-JointAtt module can be used without manual division of parts in skeleton graph, thus eliminates the need of designing appropriate pooling rules over joints in each part. Accordingly, we choose the ST-JointAtt module for building an efficient and general baseline with as few manual interventions as possible.

\begin{figure}[t]
  \centerline{\includegraphics[width=9cm]{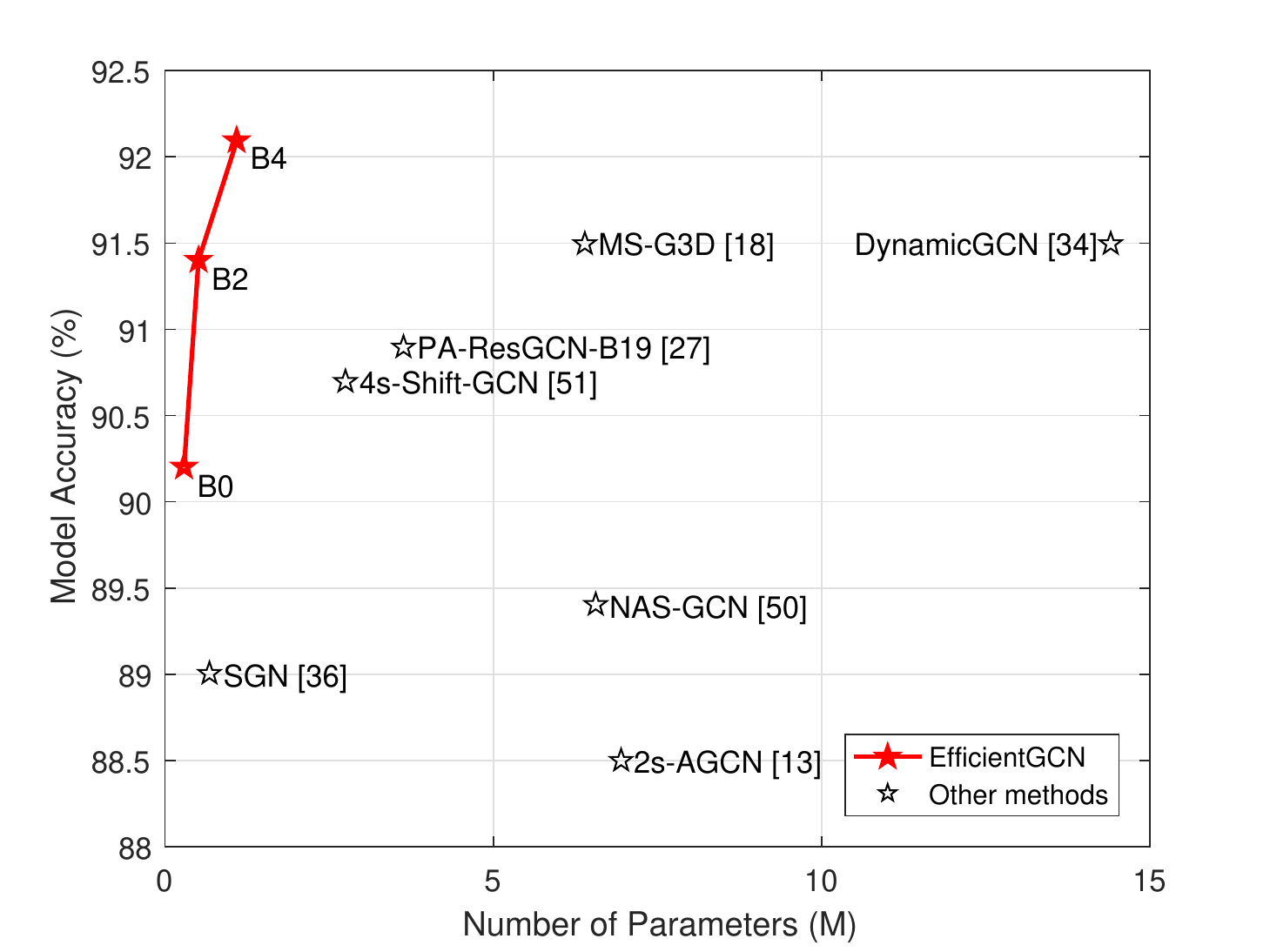}}
  \vspace{-0.4cm}
  \caption{Model size vs. model accuracy on the cross-subject benchmark of NTU 60 dataset. \bv}\label{fig:acc_param}
  \vspace{-0.4cm}
\end{figure}

Combining these efforts mentioned above, a family of efficient GCN baselines with relatively small amounts of trainable parameters while keeping competitive performance to other SOTA methods is obtained and termed EfficientGCN-Bx, where ''x'' denotes the scaling coefficient. The whole pipeline of EfficientGCN is shown in Fig.~\ref{fig:pipeline}, where the three input sequences (Joint, Velocity and Bone) are initially extracted from the original skeleton sequence. Next, each sequence is sent to a separate input branch consisting of several convolutional blocks. Then, the three branches will be fused and passed through the main stream. Similar to the input branches, the main stream is also composed of a number of convolutional blocks. Finally, the features of all frames and joints are globally averaged and fed into a Fully Connected (FC) layer for action classification. In this paper, three types of EfficientGCN with scaling coefficients of \{0,2,4\} are provided to verify the effectiveness of our approach. Compared to the most popular baseline, \ie, 2s-AGCN \cite{shi2019two}, the EfficientGCN-B0 achieves over 1\% relative performance improvement on both NTU RGB+D 60 \cite{shahroudy2016ntu} and 120 \cite{liu2019ntu} datasets, while only needs {\bf 23.93$\times$} smaller amount of model parameters and achieves {\bf 13.67$\times$} faster inference speed. Besides, EfficientGCN-B4 obtains the SOTA performance on the two datasets, \eg, {\bf 92.1\%} on the cross-subject benchmark of NTU 60 dataset. Furthermore, when considering the model size and graph computational cost, EfficientGCN-B4 is {\bf 5.82$\times$} smaller and {\bf 5.85$\times$} faster than MS-G3D \cite{liu2020disentangling}, which is one of the best SOTA methods in the field. For a clear illumination, Fig.~\ref{fig:acc_param} is drawn to demonstrate the accuracy-parameter performance of EfficientGCN, where the EfficientGCN is remarkably better than other SOTA methods.

This work is an extension of an earlier and preliminary version presented in \cite{song2020stronger}, namely ResGCN. Compared to our previous work, main modifications and contributions of this paper are summarized as follows:
\begin{itemize}
  \item For temporal convolutional layers, ResGCN proves that the BottleLayer is efficient for the GCN network. Besides, this work further introduces other three types of layers (SepLayer, EpSepLayer and SGLayer) to skeleton-based action recognition, further improving the model efficiency.
  \item In previous work, each block in the model is constructed with manually selected hyper-parameters, where the number of layers and channels are fixed. In contrast, this study employs a compound scaling strategy to configure the model's width and depth with a scaling coefficient, which brings an effective and more flexible approach to design the model architecture.
  \item In ResGCN, an attention module named PartAtt is proposed to assign spatial attentions to body parts, while this paper offers a fine-grained module (ST-JointAtt), which not only considers the spatial attention, but also distinguishes the most essential temporal frames.
  \item Compared to the preliminary version, EfficientGCN achieves a better performance with a significantly lower number of parameters and calculations on two large-scale datasets, \ie, NTU RGB+D 60 \& 120. For example, EfficientGCN-B4 obtains a {\bf 92.1\%} accuracy on the cross-subject benchmark with only 2.03 million parameters, significantly smaller and faster than ResGCN.
\end{itemize}

The remainder of this paper is organized as follows: Sec.~\ref{sec:related} describes recent studies related to our work. Sec.~\ref{sec:techniques} briefly introduces several crucial techniques of the proposed EfficientGCN. Sec.~\ref{sec:efficientgcn} presents the details of our EfficientGCN baselines. Extensive experiments on two large-scale datasets are reported in Sec.~\ref{sec:experiments}, and the conclusion is given in Sec.~\ref{sec:conclusion}.

\section{Related Work}
\label{sec:related}

\subsection{Skeleton-based Action Recognition}
\label{ssec:related_skeleton}

Due to its compactness to the RGB-based representations, action recognition based on skeleton data has received increasing attentions. In an earlier work \cite{li2018co}, a convolutional co-occurrence feature learning framework is proposed, where a hierarchical methodology is employed to gradually aggregate different levels of contextual information. The study in \cite{zhang2019view} designs a view adaptive model to automatically regulate observation viewpoints during the occurrence of an action, so as to obtain view invariant representations of human actions. However, due to the ignorance of spatial configurations, these CNN or RNN-based models gradually fade out from the stage of frontier research.

Inspired by the booming graph-based methods, Yan \etal \cite{yan2018spatial} firstly introduce GCN into the skeleton-based action recognition task, and propose the ST-GCN to model the spatial configurations and temporal dynamics of skeletons synchronously. Following this work, Song \etal \cite{song2019richly,song2020richly} aim to solve the occlusion problem in this task, and propose a multi-stream GCN to extract rich features from more activated skeleton joints. Liu \etal \cite{liu2020disentangling} explore the effects of multi-adjacency GCN and dilated temporal CNN, and design a sophisticated model named MS-G3D to disentangle multi-scale graph convolutions. Furthermore, the study in \cite{cheng2020decoupling} provides a decoupling GCN to boost the graph modeling ability with no extra computation. To achieve global joint relationship modeling, Shi \etal \cite{shi2019two} introduce the Non-local method \cite{wang2018non} into a two-stream GCN model, named 2s-AGCN, which significantly improves the recognition accuracy. Similar as 2s-AGCN, Dynamic GCN proposed by Ye \etal \cite{ye2020dynamic} offers a novel method to model global dependency, by which the model achieves outstanding accuracy for skeleton-based action recognition. Although these methods achieve considerable performance, the increasing computational cost caused by the multi-stream structure becomes the obstacle to apply them in real scenarios. Therefore, how to reduce the complexity of the GCN models is still a challenging problem.

\subsection{Efficient Models}
\label{ssec:related_efficient}

The model efficiency commonly indicated by the number of trainable parameters and Floating-point Operations Per Second (FLOPs), is always a non-negligible indicator in deep learning tasks. Extensive studies have made efforts to enhance the efficiencies of neural networks, \ie, reducing the amount of model parameters or FLOPs, such as MobileNetv1 \cite{howard2017mobilenets}, MobileNetv2 \cite{sandler2018mobilenetv2}, MobileNeXt \cite{zhou2020rethinking}, and EfficientNet \cite{tan2019efficientnet}. The model family of MobileNet mainly cuts the model size by separable convolutions, which factorizes standard convolutions into a depth-wise convolution applied to every channel individually and a $1\times1$ point-wise convolution to combine the outputs of the depth-wise convolution. To further determine the structural hyper-parameters in neural networks, compound scaling \cite{tan2019efficientnet} is proposed to build a family of EfficientNet models.

Some existing studies for skeleton-base action recognition have also been considering the model complexity problem. The study of \cite{yang2019make} constructs a lightweight network with CNN-based blocks, which is not as accurate as GCN models. The work in \cite{zhang2020semantics} adopts a complex data preprocessing strategy, whose inputs include positions, velocities, frame indexes and joint types. This data preprocessing module enables the model to recognize actions with a shallow model, thereby achieves a very fast inference speed with 188 sequences/(second*GPU), yet its performance is obviously lower than other SOTA models.

\subsection{Attention Models}
\label{ssec:related_attention}

Attention mechanisms have become an integral part of compelling sequence modeling in various tasks. Traditional attention modules for image processing can be divided into two categories: 1) channel-wise and 2) spatial-wise. Specifically, SENet \cite{hu2018squeeze} uses a bottleneck structure to obtain attention scores at channel dimension, providing a paradigm to build channel-wise attentions. Based on SENet, CBAM \cite{woo2018cbam} not only focuses on channel-wise attention, but also utilizes convolutional layers to calculate attention maps at spatial dimension for adaptive feature refinement. Along with the popularity of Self Attention (SelfAtt) in Natural Language Processing (NLP), the Non-Local \cite{wang2018non} method employs SelfAtt at spatial dimension, which globally explores attentions for the relationship between each pair of pixels.

With respect to action recognition, Baradel \etal \cite{baradel2017human} introduce the attention mechanism into an RGB-based action recognition model, which uses human poses to calculate spatial and temporal attentions. The study in \cite{song2017end} firstly introduces attention modules into skeleton-based action recognition, where a spatial-temporal attention Long Short-Term Memory (LSTM) is built to allocate different levels of attention to the discriminative joints within each frame. Si \etal \cite{si2019attention} also incorporate attention modules within LSTM units. Both of the two models apply attention modules for each frame individually, which may attend to some unstable noisy features. Besides, 2s-AGCN \cite{shi2019two} offers a variant of attention model based on the Non-Local structure, and its improved version Dynamic-GCN \cite{ye2020dynamic} proposes another way to obtain the globally spatial attentions. In addition, Cheng \etal \cite{cheng2020decoupling} embeds attention into its DropGraph module, leading to a significant accuracy increase.

\begin{figure}[t]
  \centerline{\includegraphics[width=9cm]{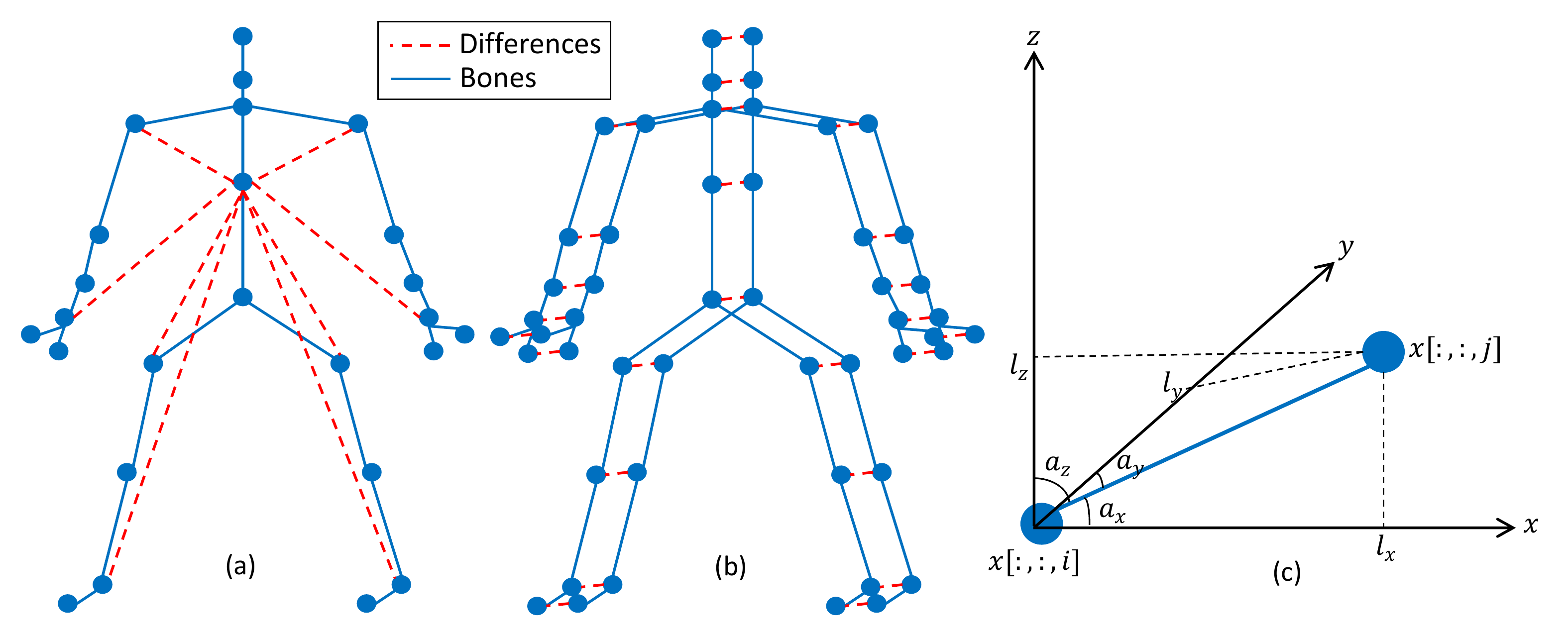}}
  \vspace{-0.4cm}
  \caption{The demonstration of input data. (a) is the relative positions, (b) is the motion velocities, and (c) demonstrates the 3D lengths and the 3D angles of a bone. \bv}\label{fig:preprocessing}
  \vspace{-0.4cm}
\end{figure}

\section{Preliminary Techniques}
\label{sec:techniques}

In this section, we briefly discuss several crucial techniques used in the proposed EfficientGCN. Firstly, the data preprocessing module is introduced and formulated. Then, the GCN layer is reviewed. Finally, we compare the separable convolution to the standard convolution.

\subsection{Data Preprocessing}
\label{ssec:preprocessing}

Data preprocessing is very essential for skeleton-based action recognition, according to previous studies \cite{song2019richly,si2018skeleton,shi2019two}. In this work, the input features after various preprocessing are mainly divided into three classes: {\bf 1)} joint positions, {\bf 2)} motion velocities and {\bf 3)} bone features.

Suppose that the original 3D coordinate set of an action sequence is $\mathcal{X}=\{x\in \mathbb{R}^{C_{in}\times T_{in}\times V_{in}}\}$, where $C_{in}$, $T_{in}$, $V_{in}$ denote the input coordinates, frames, and joints, respectively. Then the relative position set is obtained as the normalized position features, \ie, $\mathcal{R}=\{r_{i}|i=1,2,\cdots,V_{in}\}$, where
\begin{equation}
  r_i=x[:,:,i]-x[:,:,c],
\end{equation}
and $c$ represents the index of the center spine joint. Next, the input of joint positions is formed by the concatenation of $\mathcal{X}$ and $\mathcal{R}$. Moreover, it is easy to obtain the two sets of motion velocities, $\mathcal{F}=\{f_t|t=1,2,\cdots,T_{in}\}$ for fast motion and $\mathcal{S}=\{s_t|t=1,2,\cdots,T_{in}\}$ for slow motion, with the following definitions
\begin{equation}
  \begin{matrix}
    f_t=x[:,t+2,:]-x[:,t,:], \\
    s_t=x[:,t+1,:]-x[:,t,:].
  \end{matrix}
\end{equation}
And the input of motion velocities is acquired by concatenating $\mathcal{F}$ and $\mathcal{S}$ for each joint to obtain a feature vector at each time. Finally, the input of bone features consists of the bone lengths $\mathcal{L}=\{l_i|i=1,2,\cdots,V_{in}\}$ and the bone angles $\mathcal{A}=\{a_i|i=1,2,\cdots,V_{in}\}$. To obtain these two sets, the lengths and angles of each bone are calculated by
\begin{equation}
  \begin{matrix}
    l_i=x[:,:,i]-x[:,:,i_{adj}], \\
    a_{i,\mathsf{w}}=arccos(\frac{l_{i,\mathsf{w}}}{\sqrt{l_{i,\mathsf{x}}^2+l_{i,\mathsf{y}}^2+l_{i,\mathsf{z}}^2}}),
  \end{matrix}
\end{equation}
where $i_{adj}$ means the adjacent joint of the $i$-th joint, and $\mathsf{w}\in\{\mathsf{x},\mathsf{y},\mathsf{z}\}$ denotes the 3D coordinates. Fig.~\ref{fig:preprocessing} displays the calculation diagram for these three inputs.

\begin{figure}[t]
  \vspace{-0.2cm}
  \centerline{\includegraphics[width=8.5cm]{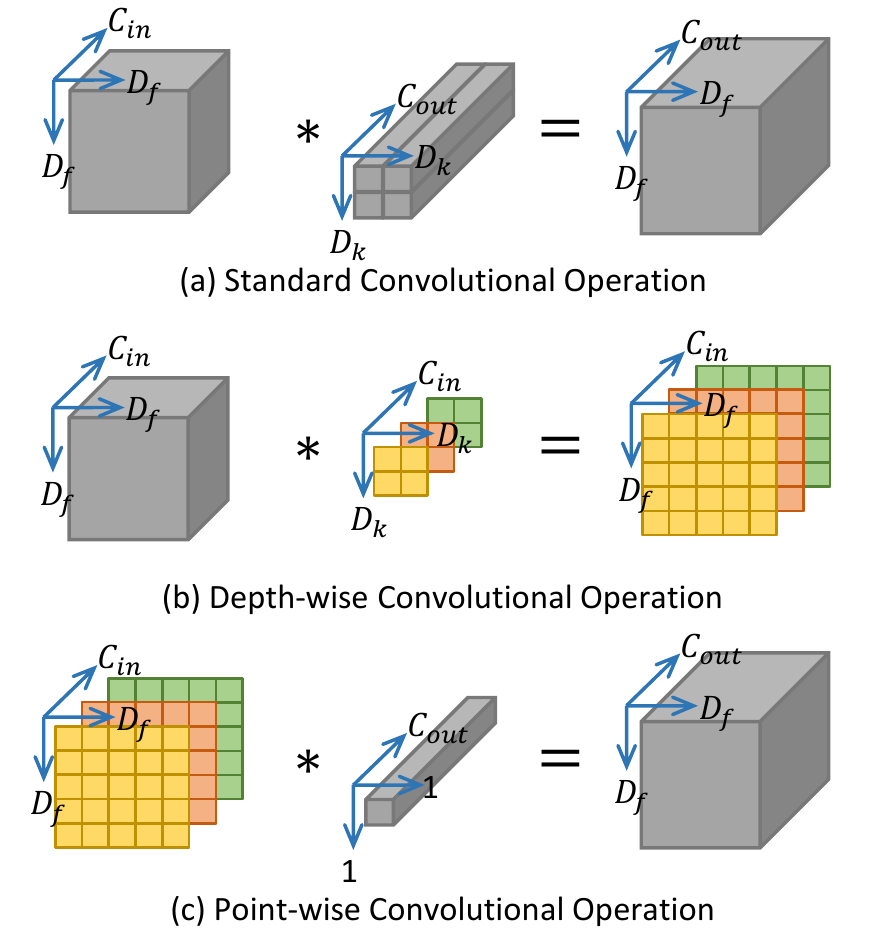}}
  \vspace{-0.4cm}
  \caption{Standard convolution vs. separable convolution for skeleton-based action recognition, where $C_{in}$ and $C_{out}$ denote the numbers of input and output channels, $D_f$ and $D_k$ denote the sizes of feature map and convolutional kernel, and $*$ represents convolutional operation.\bv}\label{fig:separable}
  \vspace{-0.4cm}
\end{figure}

\subsection{Graph Convolution}
\label{ssec:graphconv}

According to Yan \etal \cite{yan2018spatial}, graph convolutional operation for each frame $t$ can be written as 
\begin{equation}\label{eq:fout}
  f_{out}(v_{ti}) = \sum_{v_{tj}\in N(v_{ti})}\frac{1}{Z_{ti}(v_{tj})} f_{in}(v_{tj})\cdot {\bf w}(l_{ti}(v_{tj})),
\end{equation}
where $v_{ti}$ denotes the $i$-th joint at the $t$-th frame, $f_{in}(\cdot)$ and $f_{out}(\cdot)$ are the input and output features of corresponding joints, $N(v_{ti})$ is the neighbor set of $v_{ti}$, the normalizing term $Z_{ti}$ is set to balance the contributions of different neighbors, ${\bf w}(\cdot)$ is a weight function to allocate weights indexed by the label function $l_{ti}(\cdot)$, which is designed to construct several neighbor sets $N(v_{ti})$ by assigning different labels to each graph node. There are three label functions in \cite{yan2018spatial}, but we only choose the distance based partition in our model, which defines $l_{ti}(v_{tj})=d(v_{ti},v_{tj})$, where $d(v_{ti},v_{tj})$ denotes the graphic distance between $v_{ti}$ and $v_{tj}$. The joints with the same distance will form a subset and share a learnable weight function ${\bf w}(\cdot)$. Generally, with the adjacency matrix $\bf A$, Eq.~\ref{eq:fout} can be transformed into:
\begin{equation}
  {\bf f}_{out} = \sum_{d=0}^{D} {\bf W}_d {\bf f}_{in} ({\bf \Lambda}_d^{-\frac{1}{2}}{\bf A}_d{\bf \Lambda}_d^{-\frac{1}{2}} \odot {\bf M}_d),
\end{equation}
where $D$ is a predefined maximum graphic distance, ${\bf f}_{in}$ and ${\bf f}_{out}$ denote the input and output feature maps, $\odot$ means element-wise product, ${\bf A}_d$ represents the $d$-th order adjacency matrix that marks the pairs of joints with a graphic distance $d$, and ${\bf \Lambda}_d$ is used to normalize ${\bf A}_d$. ${\bf W}_d$ and ${\bf M}_d$ are both learnable parameters, which are utilized to implement the convolution operation and tune the importance of each edge, respectively.

\subsection{Separable Convolution}
\label{ssec:sepconv}

Separable convolution is initially designed as the core layers based on which MobileNet \cite{howard2017mobilenets} is built, aiming at the deployment of deep learning models on computationally limited platforms such as robotics, self-driving car, augmented reality, \etc. As its name implies, separable convolution factorizes a standard convolution into a depth-wise convolution and a point-wise convolution. Concretely, for depth-wise convolution, a convolutional filter is only applied to one corresponding channel, while the point-wise convolution uses a $1\times1$ convolution layer to combine the output of depth-wise convolution and to adjust the number of output channels. The comparison of standard convolution and separable convolution is displayed in Fig.~\ref{fig:separable}.

Concretely, suppose that the input feature size is $D_f\times D_f$, the kernel size is $D_k\times D_k$, the number of input and output channels are $C_{in}$ and $C_{out}$, then the calculational process of standard convolution is illustrated in the top row of Fig.~\ref{fig:separable}. This brings a batch of trainable parameters numbered $D_k\times D_k\times C_{in}\times C_{out}\times D_f\times D_f$. With respect to separable convolution shown in the bottom two rows of Fig.~\ref{fig:separable}, the computational cost is changed to $D_k\times D_k\times C_{in}\times D_f\times D_f + C_{in}\times C_{out}\times D_f\times D_f$. Note that the most of trainable parameters are contained in the point-wise convolution \cite{howard2017mobilenets}, which is implemented by a $1\times1$ convolution. Thus, if the numbers of input and output channels are big enough, \eg, $>256$, then the computational cost of separable computation will be decreased by nearly $D_k\times D_k$ times compared to that of standard convolution.

\begin{figure}[t]
  \vspace{-0.2cm}
  \centerline{\includegraphics[width=8.5cm]{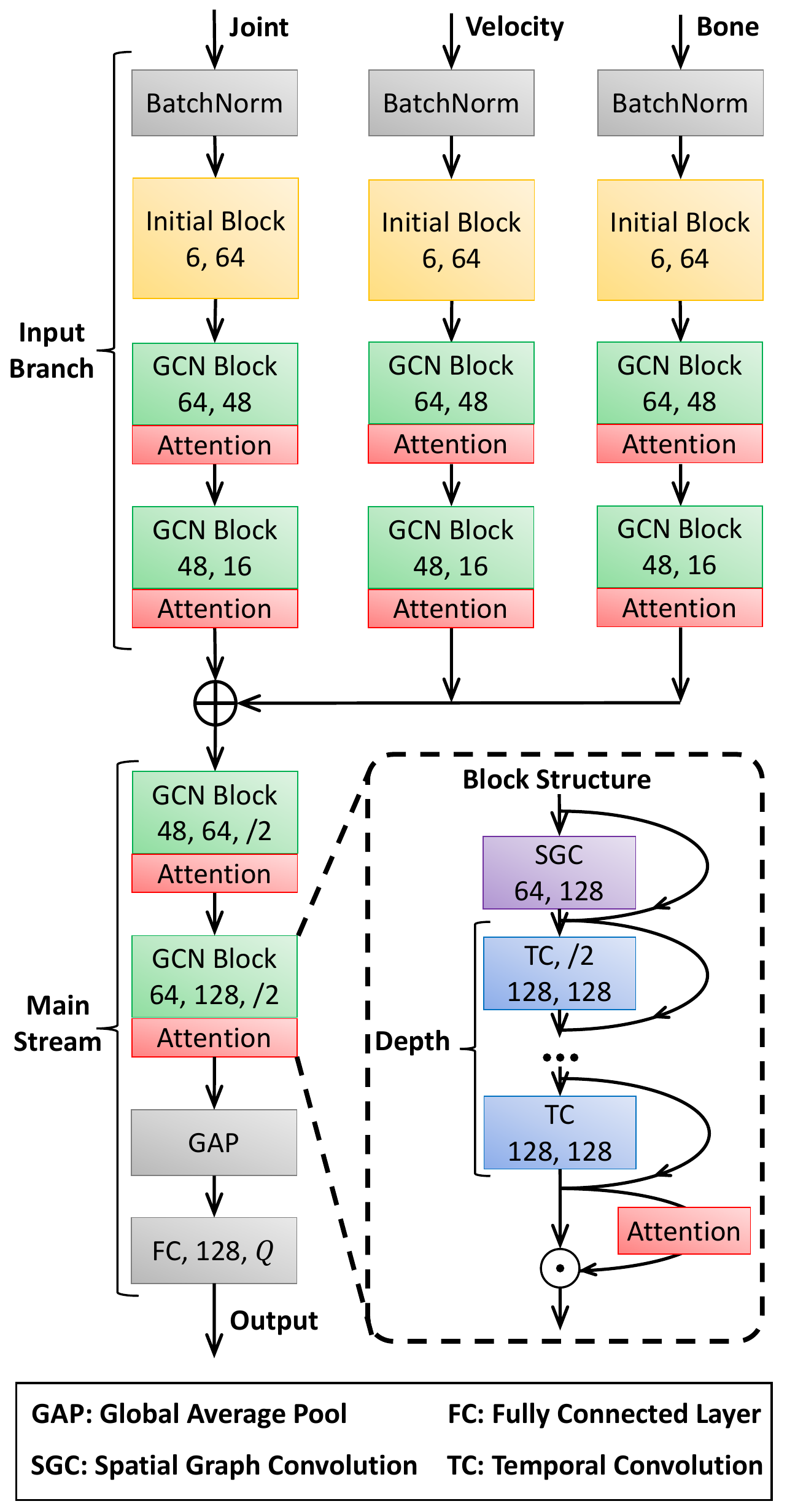}}
  \vspace{-0.4cm}
  \caption{The overview of the proposed EfficientGCN model, where the two numbers in each block denote input and output channels, $Q$ is the number of action classes, $\oplus$ and $\odot$ represent concatenation and element-wise product, and $/2$ represents a stride of 2. \bv}\label{fig:model}
  \vspace{-0.4cm}
\end{figure}

\begin{figure*}[t]
  \centerline{\includegraphics[width=17cm]{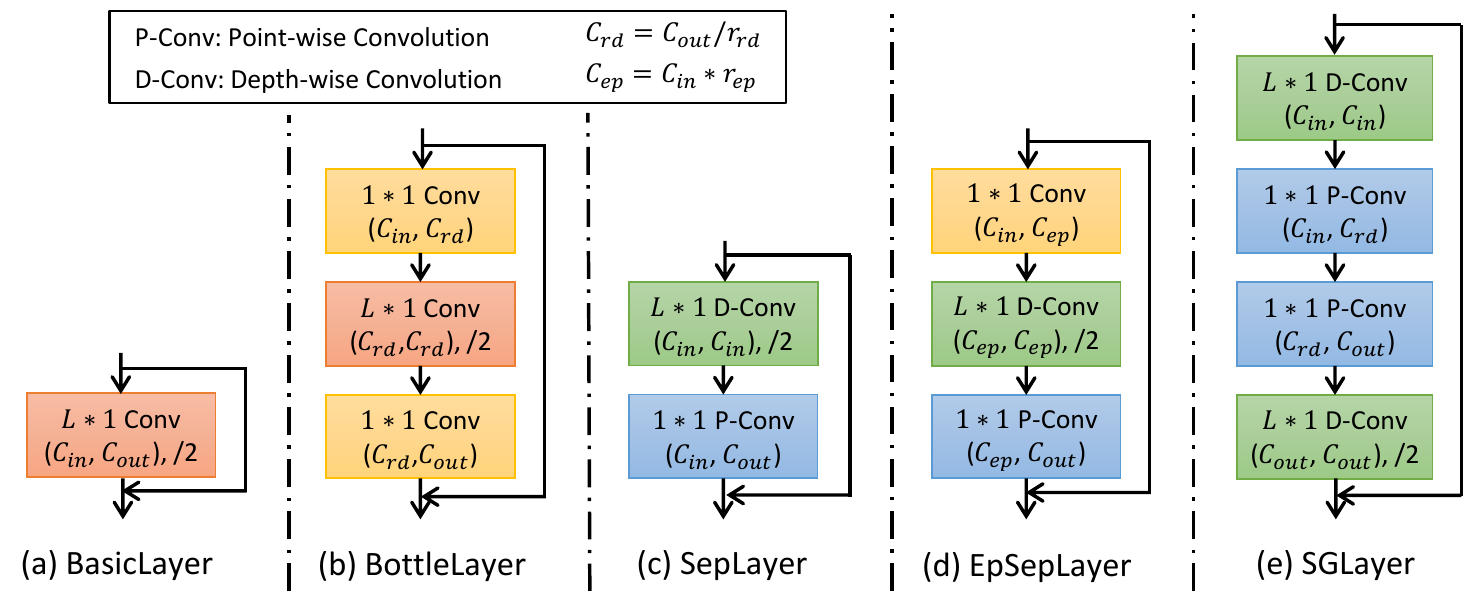}}
  \vspace{-0.4cm}
  \caption{The details of various convolutional layers, where $C_{in}$ and $C_{out}$ denote the numbers of input and output channels, $r_{rd}$ and $r_{ep}$ are employed to reduce or expand the inner channels. \bv}\label{fig:layers}
  \vspace{-0.4cm}
\end{figure*}

\section{EfficientGCN}
\label{sec:efficientgcn}

This part provides technical details to the proposed EfficientGCN for skeleton-based action recognition. Firstly, the MIB architecture is discussed and an example of EfficientGCN-B0 is constructed. Then, four kinds of convolutional layers popularly used in CNNs are extended to graph convolution to increase the efficiency of GCN blocks. Next, a compound scaling strategy is utilized to synchronously scale the width and depth of EfficientGCN-B0, generating a family of efficient baselines. Finally, an attention module is proposed to enhance the discrimination of skeleton features.

\subsection{Model Architecture}
\label{ssec:architecture}

After the data preprocessing module designed in Sec.~\ref{ssec:preprocessing}, three types of input data are obtained, \ie, Joint, Velocity and Bone. For current high-performance complex models, they usually apply a multi-stream architecture to handle these input data. For example, Shi \etal \cite{shi2019two} take the joint data and bone data as inputs for feeding to two GCN branches with the same model structures separately, and eventually choose the fusion results of two streams as the final decision. This is an effective way to augment the input data and enhance the model performance. However, a multi-stream network often means high computational cost and difficulties of parameter turning on large-scale datasets. Thus, we devise the MIB architecture that fuses the three input branches at the early stage of our model, then apply one main stream to extract discriminative features. This architecture not only retains the rich input features, but also significantly suppresses the model complexity with fewer parameters, thus is easier to be trained. An example of EfficientGCN with the MIB architecture is demonstrated in Fig.~\ref{fig:model}.

Concretely, the input branches are formed by orderly stacking a BatchNorm layer for fast convergence, an initial block implemented by ST-GCN layer \cite{yan2018spatial} for data-to-feature transformation, and two GCN blocks with attentions for informative feature extraction. After the input branches, a concatenation operation is employed to fuse the feature maps of three branches and then send them to the main stream, which is constructed by two GCN blocks. Finally, the output feature map of the main stream is globally averaged to a feature vector, and an FC layer is used to determine the final action class.

\subsection{Block Details}
\label{ssec:block}

Inspired by MS-G3D \cite{liu2020disentangling} which achieves a considerable performance, as the subplot of Fig.~\ref{fig:model} shows, the basic components of EfficientGCN (\ie, GCN blocks) are implemented by orderly stacking a Spatial Graph Convolutional (SGC) layer, several Temporal Convolutional (TC) layers and an attention module. The depth for each GCN block is defined as the number of TC layers stacked in this block. Besides, for each layer, a residual link is utilized to make the model optimization more easily than the original unreferenced feature projection. It also should be noted that the first TC layer has a stride of 2 for each block in the main stream, which aims to compress the features and reduce the convolutional costs.

\begin{figure*}[t]
  \centerline{\includegraphics[width=17cm]{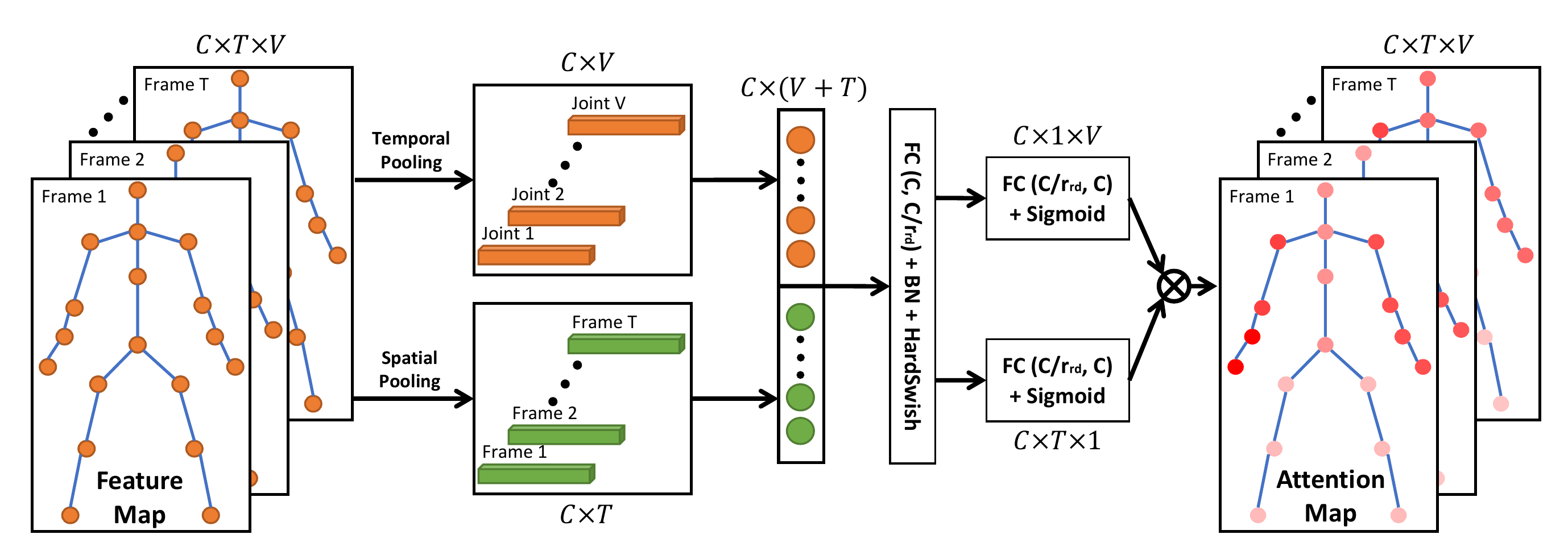}}
  \vspace{-0.4cm}
  \caption{The overview of the proposed ST-JointAtt module, where $C,T,V$ denote the numbers of input channels, frames and joints respectively, $r_{rd}=4$ is utilized to compact the features, $\otimes$ represents the outer-product, BN denotes the batch normalization, HardSwish \cite{howard2019searching} and Sigmoid are both activated functions. \bv}\label{fig:stjatt}
  \vspace{-0.4cm}
\end{figure*}

In detail, the SGC layer is implemented by a graph convolution mentioned in Sec.~\ref{ssec:graphconv}, and the attention module can be implemented by the proposed ST-JointAtt (see Sec.~\ref{ssec:stjatt}) or other traditional attention modules. For the implementation of TC layer, except for the basic $L\times1$ convolution ({\bf BasicLayer}) originally used in ST-GCN model \cite{yan2018spatial}, we introduce four types of convolutional architectures widely used in CNN literatures to further boost the efficiency of GCN models (see Fig.~\ref{fig:layers}). Specifically, for {\bf BottleLayer}, a bottleneck structure \cite{he2016deep} is utilized for temporal convolution, which is also used in our preliminary version of this paper, \ie, ResGCN \cite{song2020stronger}. The other three layers, \ie, {\bf SepLayer}, {\bf EpSepLayer} and {\bf SGLayer}, are inspired by three versions of separable convolutions \cite{howard2017mobilenets,sandler2018mobilenetv2,zhou2020rethinking}, all of which are composed of depth-wise convolutions and point-wise convolutions mentioned in Sec.~\ref{ssec:sepconv}. Note that the block with a certain layer, \eg, BasicLayer, is denoted as {\bf BasicBlock} for simplicity, and by analogy to {\bf BottleBlock}, {\bf SepBlock}, {\bf EpSepBlcok} and {\bf SGBlock}.

\subsection{Scaling Strategy}
\label{ssec:scaling}

Empirically, expanding both the width (\eg, WRN \cite{zagoruyko2016wide}) and the depth (\eg, ResNet \cite{he2016deep}) of networks will benefit the model capability and hence improve the performance. Commonly, the model's width and depth are defined as the numbers of channels and layers. These two factors are often considered independently, and determined by handcrafted adjustment. However, in a recent work \etal \cite{tan2019efficientnet}, Tan and Le show that it is critical to balance all dimensions of network, \eg, width/depth/resolution, based on which a compound scaling strategy is proposed to scale network width, depth, and resolution with a set of fixed scaling coefficients. The obtained EfficientNets significantly outperform other convolutional networks, but with much smaller model sizes. Inspired by that, after removing the resolution factor, we propose a new scaling strategy for skeleton-based action recognition, by which a family of models are constructed in a principled way:
\begin{equation}\label{eq:scaling}
  \begin{array}{rl}
    \vspace{0.2cm}
    \text{width:} & m_w=\alpha^\phi \\
    \vspace{0.2cm}
    \text{depth:} & m_d=\beta^\phi \\
    \vspace{0.2cm}
    \text{s.t.}   & \alpha^2\cdot\beta\approx 2 \\
                  & \alpha\geq1,\beta\geq1 \\
  \end{array}
\end{equation}
where $m_w$ and $m_d$ are width and depth multipliers, $\phi$ is a compound coefficient to state the available resources for model scaling, $\alpha$ and $\beta$ are both hyper-parameters to control the resource assignments to the model's width and depth. In this paper, $\alpha$ and $\beta$ are set to 1.2 and 1.35 by a small grid search (see Sec.~\ref{ssec:compound}). Here, the resolution factor is omitted due to the pre-defined skeleton structure. The modification of skeleton's resolution will destroy the graph convolutional operation.

The constraint conditions in Eq.~\ref{eq:scaling} are utilized to constrain the increasing speed of the model size. When doubling the model depth or width, the FLOPs will approximately increase to 2 or 4 times. Thus, the FLOPs of the scaled model will be $(\alpha^2\cdot\beta)^\phi \approx 2^\phi$ times than the baseline. Note that this increase may differ from theoretical values due to the rounding function (see Appendix~\ref{asec:architectures}).

\subsection{Spatial Temporal Joint Attention}
\label{ssec:stjatt}

Previous attention modules for skeleton-based action recognition are mainly implemented by a Multi-layer Perception (MLP) like SENet structure \cite{hu2018squeeze}, such as AGC-LSTM \cite{si2019attention} and MS-AAGCN \cite{shi2020skeleton}. These modules are usually performed on each channel or spatial dimension independently, while other dimensions are globally averaged to a single unit. The preliminary version of this paper \cite{song2020stronger} proposes a PartAtt module which only works on the spatial dimension. However, intuitively, the spatial and temporal information could be relevant to each other. Thus, separately considering frames and joints is sub-optimal for weighting the importance of skeleton joints in different action phases. To address this issue, inspired by coordinate attention \cite{hou2021coordinate}, we propose a novel attention module, named Spatial Temporal Joint Attention (ST-JointAtt), to jointly distinguish the most informative joints in certain frames from the whole skeleton sequence.

The overview of the proposed ST-JointAtt module is shown in Fig.~\ref{fig:stjatt}, from which the input features are firstly averaged in frame- and joint-level respectively. Then, these pooled feature vectors are concatenated together and fed through an FC layer to compact information. Next, two independent FC layers are utilized to obtain two sets of attention scores for frame dimension and joint dimension respectively. Finally, the scores of frames and joints are multiplied by channel-wise outer-product, and the result can be seen as the attention scores for the whole action sequence. The proposed ST-JointAtt module can be formulated as
\begin{equation}
  f_{inner}=\theta((pool_t(f_{in}) \oplus pool_v(f_{in}))\cdot W)
\end{equation}
\begin{equation}
  f_{out}=f_{in}\odot(\sigma(f_{inner}\cdot W_t)\otimes\sigma(f_{inner}\cdot W_v))
\end{equation}
where $f_{in}$ and $f_{out}$ denote input and output feature maps, $\oplus$ denotes concatenation operation, $\otimes$ and $\odot$ mean channel-wise outer-product and element-wise product, $pool_t(\cdot)$ and $pool_v(\cdot)$ are average pooling operations on frame- and joint-level respectively, $\sigma(\cdot)$ and $\theta(\cdot)$ represent Sigmoid and HardSwish  \cite{howard2019searching} activation functions, and $W\in\mathbb{R}^{C\times\frac{C}{r}}$, $W_t\in\mathbb{R}^{\frac{C}{r}\times C}$, $W_v\in\mathbb{R}^{\frac{C}{r}\times C}$ are trainable parameters.

\subsection{Loss Function}
\label{ssec:loss}

Suppose that the final FC layer in the EfficientGCN model outputs a logits vector ${\bf z}\in\mathbb{R}^{Q}$, where $Q$ is the number of action classes, the classification probabilities for the input sample can be computed by a Softmax function, \ie,
\begin{equation}
  \hat{y}_i=\frac{e^{z_i}}{\sum_{j=1}^{Q}e^{z_j}},i=1,2,\cdots,Q,
\end{equation}
where $z_i$ denotes the $i$-th element of ${\bf z}$. Then, a cross-entropy loss is calculated as the objective function for model optimization:
\begin{equation}
L=-\sum_{i=1}^{Q}y_i\log\hat{y}_i
\end{equation}
where ${\bf y}\in\mathbb{R}^{Q}$ is the one-hot vector indicating the ground truth of action class.

\subsection{Discussion}
\label{ssec:discussion}

In this section, we interpret why the EfficientGCN method can achieve superior accuracy but with much fewer model parameters than traditional GCN models. Firstly, the use of separable convolution in TC layers brings high efficiency to the model, by which the EfficientGCN reduces model parameters and FLOPs significantly. It should be noticed that the parameter reduction caused by separable convolution may not always hurt the model performance, since separable convolution has been proved to capture the most effective part of standard convolutions and meanwhile discard other redundant parts \cite{guo2018network}.

Moreover, as to the superior accuracy achieved by the EfficientGCN, it is mainly attributed to the compound scaling strategy and the ST-JointAtt module, where the former one is based on an empirical experience in CNN-based visual recognition that carefully balancing network depth, width, and resolution can lead to better performance \cite{tan2019efficientnet}, and the latter one further enhances the learning of spatial-temporal joint features through making the GCN attends on those informative joints and frames in action sequences.

To validate the above analysis, extensive ablation studies have been performed to evaluate the impacts and contributions of the above three components in the next section (see Section Sec.~\ref{ssec:ablation}, \ref{ssec:compound}, and \ref{ssec:comparisons}).

\section{Experimental Results}
\label{sec:experiments}

In this section, we evaluate the proposed EfficientGCN on two large-scale datasets, \ie, NTU RGB+D 60 \cite{shahroudy2016ntu} and NTU RGB+D 120 \cite{liu2019ntu}. Ablation studies are also performed to validate the contribution of each component in our model. For simplicity, all experiments in ablation studies choose EfficientGCN-B0 with SGBlock ($r_{rd}=2$) as the baseline model (details can be seen in Appendix~\ref{asec:architectures}). Finally, result analysis and visualization are reported to prove the effectiveness of the proposed method.

\subsection{Datasets}
\label{ssec:datasets}

\subsubsection{NTU RGB+D 60}
\label{sssec:dataset_ntu60}

This large-scale indoor dataset is provided in \cite{shahroudy2016ntu}, which contains 56880 human action videos collected by three Kinect v2 cameras. These actions consist of 60 classes, where the last 10 classes are all interactions between two subjects. For simplicity, the input frame number is set to 300, and the sequences with less than 300 frames are padded by 0 at the end. Each frame contains no more than 2 skeletons, and each skeleton is composed of 25 joints. The authors of this dataset recommend two benchmarks: {\bf 1) cross-subject (X-sub)} contains 40320 training videos and 16560 evaluation videos divided by splitting the 40 subjects into two groups. {\bf 2) cross-view (X-view)} recognizes the videos collected by cameras 2 and 3 as training samples (37920 videos), while the videos collected by camera 1 are treated as evaluation samples (18960 videos). Note that there are 302 wrong samples selected by \cite{liu2019ntu} that need to be ignored.

\subsubsection{NTU RGB+D 120}
\label{sssec:dataset_ntu120}

This is currently the largest indoor skeleton-based action recognition dataset \cite{liu2019ntu}, which is an extended version of the NTU RGB+D 60. It totally contains 114480 videos performed by 106 subjects from 155 viewpoints. These videos consist of 120 classes, extended from the 60 classes of the previous dataset. Similarly, two benchmarks are suggested: {\bf 1) cross-subject (X-sub120)} divides subjects into two groups, to construct training and evaluation sets (63026 and 50922 videos respectively). {\bf 2) cross-setup (X-set120)} contains 54471 videos for training and 59477 videos for evaluation, which are separated based on the distance and height of their collectors. According to \cite{liu2019ntu}, 532 bad samples of this dataset should be ignored in all experiments.

\subsection{Implementation Details}
\label{ssec:implementation}

In our experiments, the maximum number of training epochs is set to 70. The initial learning rate is set to 0.1 and decays with a cosine schedule after the 10th epoch. Moreover, a warmup strategy \cite{he2016deep} is applied over the first 10 epochs, gradually increasing the learning rate from 0 to the initial value for a stable training procedure. The stochastic gradient descent (SGD) with the Nesterov momentum of 0.9 and the weight decay of 0.0001 is employed to tune the parameters. The hyper-parameters $D$ and $L$ defined in Sec.~\ref{ssec:graphconv} and \ref{ssec:block} are set to 2 and 5 respectively, which are determined by a grid search (see Appendix~\ref{asec:receptive}). Other structural parameters will be discussed in ablation studies (Sec.~\ref{ssec:ablation}). In addition, a dropout layer with 0.25 drop probability is added after the GAP layer and before the final FC layer to avoid overfitting. It also should be noticed that the activated function used in all convolutional blocks is chosen as Swish function \cite{ramachandran2017searching}, which is similar with ReLU function but smooth and differentiable everywhere. In our experiments on X-view benchmark, a special data transformation \cite{shi2019two} is performed for view alignment. All our experiments are performed on two TITAN RTX GPUs.

\subsection{Ablation Studies}
\label{ssec:ablation}

In this part, we mainly discuss the contributions of different components in the proposed EfficientGCN, including the selection of TC layers, the choice of the attention modules, the importance of data preprocessing module, and the necessity of the early fused architecture. This section explains why we use these settings to construct the baseline model EfficientGCN-B0. All of the experiments in this section are performed more than 10 times to compute the mean accuracy and standard error for convincing results.

\begin{table}[t]
  \caption{Comparisons of different TC layer types on X-sub benchmark in accuracy (\%), FLOPs ($\times10^9$) and parameter number ($\times10^6$).}
  \label{tab:layer}
  \vspace{-0.4cm}
  \centering
  \setlength{\tabcolsep}{4pt}
  \renewcommand{\arraystretch}{1.2}
  \begin{tabular}{c|ccc}
  \toprule
  Layer & Mean$\pm$Std. & FLOPs & \# Param.\\
  \midrule
  BasicLayer & 90.0$\pm$0.12 & 2.96 & 0.34 \\
  \midrule
  BottleLayer ($r_{rd}=4$) & 89.6$\pm$0.15 & 2.62 & 0.26 \\
  \midrule
  SepLayer & 89.6$\pm$0.15 & 2.62 & 0.26 \\
  \midrule
  EpSepLayer ($r_{ep}=1$) & 89.6$\pm$0.21 & 2.80 & 0.28 \\
  EpSepLayer ($r_{ep}=2$) & 89.9$\pm$0.19 & 3.08 & 0.32 \\
  EpSepLayer ($r_{ep}=4$) & 90.1$\pm$0.15 & 3.63 & 0.41 \\
  \midrule
  SGLayer ($r_{rd}=2$) (Baseline) & {\bf 90.0$\pm$0.10} & {\bf 2.73} & {\bf 0.29} \\
  SGLayer ($r_{rd}=4$) & 89.8$\pm$0.13 & 2.63 & 0.25 \\
  \bottomrule
  \end{tabular}
\end{table}

\subsubsection{Comparisons of TC Layers}
\label{sssec:compare_layer}

In Sec.~\ref{ssec:block} and Fig.~\ref{fig:layers}, five types of TC layers are provided, namely BasicLayer, BottleLayer, SepLayer, EpSepLayer and SGLayer. To select the best TC layer for skeleton-based action recognition, we test them with the EfficientGCN-B0 model on X-sub benchmark, and the results are presented in Tab.~\ref{tab:layer}, where $r_{rd}$ and $r_{ep}$ denote the ratios of reducing and expanding channels in the corresponding layer. As shown in Tab.~\ref{tab:layer}, although the EpSepLayer ($r_{ep}=4$) obtains the highest mean accuracy (90.1\%), the SGLayer with $r_{rd}=2$ achieves the optimal trade-off between performance and computational cost, resulting in the competitive mean accuracy of 90.0\%, lower standard derivation of 0.10\%, faster inference speed with 2.73G FLOPs, and smaller model size with 0.29M parameters. Therefore, we choose the SGLayer with $r_{rd}=2$ to build the baseline model.

\subsubsection{Comparisons of Attention Modules}
\label{sssec:compare_attention}

To enhance the model ability to extract informative features, the ST-JointAtt module is designed and embedded into the convolutional blocks. We compare ST-JointAtt with other attention modules, \ie, ChannelAtt, FrameAtt, JointAtt, STCAtt, and PartAtt, and the results are presented in Tab.~\ref{tab:attention}, where the first three attention modules are designed by ourselves based on the SENet \cite{hu2018squeeze} structure, and the last two are proposed in \cite{shi2020skeleton,song2020stronger}, respectively. It is observed that, after incorporating attention modules, the model achieves obvious accuracy improvement. And ST-JointAtt produces the best accuracy, while other five modules are slightly worse. This indicates the importance of inserting attention modules and the effectiveness of the ST-JointAtt module in finding the most informative joints and frames from the whole skeleton sequence.

\begin{table}[t]
  \caption{Comparisons of different attention modules on X-sub benchmark in accuracy (\%), FLOPs ($\times10^9$) and parameter number ($\times10^6$).}
  \label{tab:attention}
  \vspace{-0.4cm}
  \centering
  \setlength{\tabcolsep}{4pt}
  \renewcommand{\arraystretch}{1.2}
  \begin{tabular}{c|ccc}
  \toprule
  Attention & Mean$\pm$Std. & FLOPs & \# Param. \\
  \midrule
  w/o Att & 88.9$\pm$0.16 & 2.72 & 0.24 \\
  w/ ChannelAtt & 89.3$\pm$0.18 & 2.72 & 0.25 \\
  w/ FrameAtt & 88.5$\pm$0.19 & 2.72 & 0.24 \\
  w/ JointAtt & 89.1$\pm$0.18 & 2.72 & 0.24 \\
  w/ STCAtt \cite{shi2020skeleton} & 89.5$\pm$0.14 & 2.74 & 0.30 \\
  w/ PartAtt \cite{song2020stronger} & 89.4$\pm$0.21 & 2.72 & 0.33 \\
  w/ ST-JointAtt (Baseline) & {\bf 90.0$\pm$0.10} & 2.73 & 0.29 \\
  \bottomrule
  \end{tabular}
\end{table}

\begin{table}[t]
  \caption{Comparisons of different inputs on X-sub benchmark in accuracy (\%), FLOPs ($\times10^9$) and parameter number ($\times10^6$).}
  \label{tab:input}
  \vspace{-0.4cm}
  \centering
  \setlength{\tabcolsep}{4pt}
  \renewcommand{\arraystretch}{1.2}
  \begin{tabular}{c|ccc}
  \toprule
  Inputs & Mean$\pm$Std. & FLOPs & \# Param. \\
  \midrule
  Joint & 87.7$\pm$0.15 & 1.28 & 0.17 \\
  Velocity & 86.6$\pm$0.21 & 1.28 & 0.17 \\
  Bone & 88.4$\pm$0.14 & 1.28 & 0.17 \\
  \midrule
  Joint + Velocity & 89.4$\pm$0.22 & 1.94 & 0.23 \\
  Joint + Bone & 88.9$\pm$0.22 & 1.94 & 0.23 \\
  Velocity + Bone & 89.7$\pm$0.13 & 1.94 & 0.23 \\
  \midrule
  Joint + Velocity + Bone (Baseline) & {\bf 90.0$\pm$0.10} & 2.73 & 0.29 \\
  \bottomrule
  \end{tabular}
\end{table}

\begin{table}[t]
  \caption{Comparisons of different fusion stages on X-sub benchmark in accuracy (\%), FLOPs ($\times10^9$) and parameter number ($\times10^6$).}
  \label{tab:fusion}
  \vspace{-0.4cm}
  \centering
  \setlength{\tabcolsep}{4pt}
  \renewcommand{\arraystretch}{1.2}
  \begin{tabular}{c|ccc}
  \toprule
  Fusion & Mean$\pm$Std. & FLOPs & \# Param. \\
  \midrule
  Before 1st stage & 88.8$\pm$0.30 & 2.29 & 0.24 \\
  After 1st stage & 89.7$\pm$0.13 & 2.48 & 0.27 \\
  After 2nd stage (Baseline) & {\bf 90.0$\pm$0.10} & 2.73 & 0.29 \\
  After 3rd stage & 89.8$\pm$0.25 & 3.43 & 0.38 \\
  After 4th stage & 89.5$\pm$0.22 & 3.84 & 0.52 \\
  At the score layer & 89.9$\pm$0.06 & 3.85 & 0.52 \\
  \bottomrule
  \end{tabular}
  \vspace{-0.4cm}
\end{table}

\subsubsection{Necessity of Data Preprocessing}
\label{sssec:compare_input}

Data preprocessing is essential to enhance the model performance, proven by several previous studies \cite{song2019richly,si2018skeleton,shi2019two}. We summarize these preprocessing methods into three classes, \ie, joint positions, motion velocities and bone features. To explore the necessity of each input branch, we present Tab.~\ref{tab:input}, from which the model with three inputs is clearly better than others. With the increase of branches, the model performance is improved. This phenomenon further confirms the effectiveness of the data preprocessing module.

\begin{figure}[t]
  \centerline{\includegraphics[width=9cm]{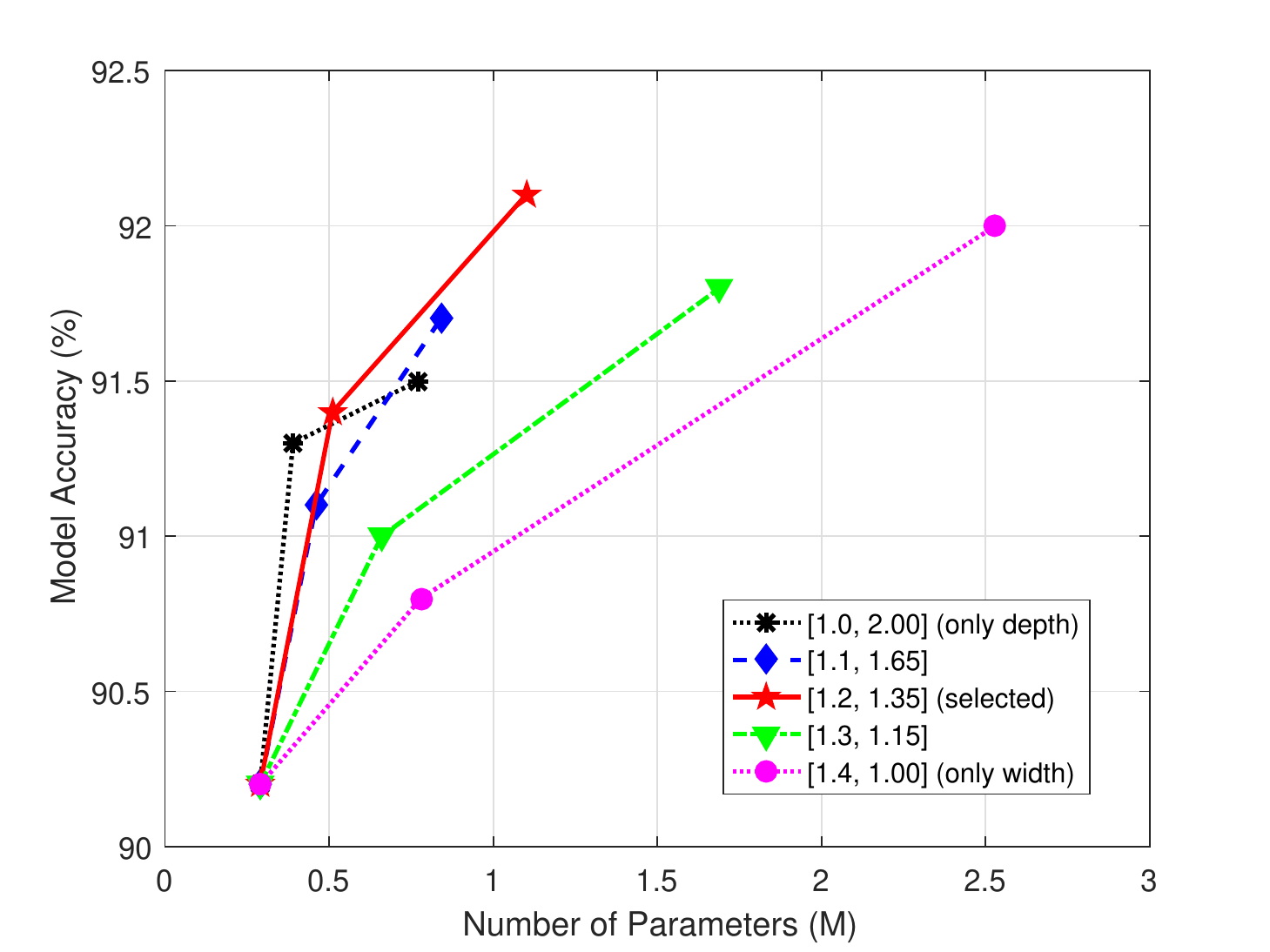}}
  \vspace{-0.4cm}
  \caption{Model size vs. model accuracy of different width and depth scaling hyper-parameters ($\alpha$ and $\beta$). \bv}\label{fig:scaling}
\end{figure}

\subsubsection{Necessity of Early Fused Architecture}
\label{sssec:compare_fusion}

Originally proposed by 2s-AGCN \cite{shi2019two}, multi-stream models fused at the final score layer gradually dominate the model architectures for skeleton-based action recognition. However, the late fusion strategy at score layer requires training three models independently to deal with three types of input data. So, this strategy cannot be implemented within an end-to-end training pipeline and thereby increases the training complexity. Furthermore, we find that many parameters in multi-stream models are redundant, with no contribution to model performance. Thus, we construct an early fused MIB architecture in this paper. Tab.~\ref{tab:fusion} gives the results of models with different fusion stages. It is seen that fusing after the 2nd stage is the inflection point of the accuracy-parameter curve, which is the best model to balance accuracy and complexity. Although fusing at the score layer brings a comparable model accuracy, the model size and computational cost are greatly increased, leading to an exceedingly sophisticated and over-parameterized model. Accordingly, the MIB fused after the 2nd stage is chosen as the architecture of our baseline model.

\subsection{Comparisons of Compound Scaling Strategies}
\label{ssec:compound}

In this section, we test several compound scaling strategies with different width and depth scaling hyper-parameters ($\alpha$ and $\beta$), which are mentioned in Sec.~\ref{ssec:scaling}. To choose the best setting of $\alpha$ and $\beta$ within the constraint conditions in Eq.~\ref{eq:scaling}, we set $\alpha$ to $\{1.0, 1.1, 1.2, 1.3, 1.4\}$, respectively, and the hyper-parameter $\beta$ can be calculated by Eq.~\ref{eq:scaling} with a precision of 0.05. Then, from the baseline model EfficientGCN-B0, we construct five EfficientGCN-B2 and five EfficientGCN-B4 with different scaling hyper-parameters. These models' accuracies and parameter numbers are shown in Fig.~\ref{fig:scaling}, from which the EfficientGCN-B4 obtains the best performance when $\alpha=1.2$ and $\beta=1.35$. It is also worth noting that, merely increasing either width or depth may hurt the model performance, whose accuracies with EfficientGCN-B4 are obviously lower than those of compound scaling strategies.

\subsection{Comparisons with SOTA Methods}
\label{ssec:comparisons}

\subsubsection{NTU RGB+D 60 Dataset}
\label{sssec:ntu60}

We compare our EfficientGCN family against previous SOTA methods on both X-sub and X-view benchmarks of NTU 60 datasets. The results are displayed in Tab.~\ref{tab:ntu60}. It should be noted that most previous studies in action recognition only report the best accuracies in comparisons. We also report the best accuracy for each EfficientGCN model for fair comparisons. There are several typical comparisons shown as follows:

From Tab.~\ref{tab:ntu60}, the best performance of the baseline model EfficientGCN-B0 are 90.2\% and 94.9\% for X-sub and X-view benchmark, respectively. When applying the proposed scaling strategy with coefficients of 2 and 4, the EfficientGCN-B2 and EfficientGCN-B4 are built, and achieve better performances on the two benchmarks. Especially for EfficientGCN-B4, its accuracy on X-sub benchmark 92.1\%, outperforming other SOTA methods. Here, three typical methods should be noticed. {\bf 1)} The first one is ST-GCN \cite{yan2018spatial}, which is currently the most popular backbone model for skeleton-based action recognition. Compared to ST-GCN, our EfficientGCN-B4 leads over 10\% on X-sub benchmark. {\bf 2)} 2s-AGCN \cite{shi2019two} is another popular baseline for skeleton-based action recognition. The proposed baseline EfficientGCN-B0 outperforms 2s-AGCN in both accuracy and efficiency. {\bf 3)} The third one is MST-GCN \cite{chen2021multi}, which is the current SOTA method with the GCN technique. The EfficientGCN-B4 is slightly better than MST-GCN in accuracy on X-sub benchmark. With respect to the comparisons within the EfficientGCN family, the accuracy shows a gradually improving trend with the increase of scaling coefficient.

\begin{table}[t]
  \caption{Comparisons with SOTA methods on NTU 60 dataset in accuracy (\%). The top part consists of several models without GCN techniques, while the other part contains some graph-based models.}
  \label{tab:ntu60}
  \vspace{-0.4cm}
  \centering
  \setlength{\tabcolsep}{4pt}
  \renewcommand{\arraystretch}{1.2}
  \begin{tabular}{cc|cc}
  \toprule
  Model & Conference & X-sub & X-view \\
  \midrule
  HBRNN \cite{du2015hierarchical} & CVPR15 & 59.1 & 64.0 \\
  ST-LSTM \cite{liu2016spatio} & ECCV16 & 69.2 & 77.7 \\
  STA-LSTM \cite{song2017end} & AAAI17 & 73.4 & 81.2 \\
  HCN \cite{li2018co} & IJCAI18 & 86.5 & 91.1 \\
  VA-fusion \cite{zhang2019view} & TPAMI19 & 89.4 & 95.0 \\
  \midrule
  ST-GCN \cite{yan2018spatial} & AAAI18 & 81.5 & 88.3 \\
  SR-TSL \cite{si2018skeleton} & ECCV18 & 84.8 & 92.4 \\
  RA-GCNv1 \cite{song2019richly} & ICIP19 & 85.9 & 93.5 \\
  RA-GCNv2 \cite{song2020richly} & TCSVT20 & 87.3 & 93.6 \\
  AS-GCN \cite{li2019actional} & CVPR19 & 86.8 & 94.2 \\
  2s-AGCN \cite{shi2019two} & CVPR19 & 88.5 & 95.1 \\
  AGC-LSTM \cite{si2019attention} & CVPR19 & 89.2 & 95.0 \\
  DGNN \cite{shi2019skeleton} & CVPR19 & 89.9 & 96.1 \\
  PL-GCN \cite{huang2020part} & AAAI20 & 89.2 & 95.0 \\
  NAS-GCN \cite{peng2020learning} & AAAI20 & 89.4 & 95.7 \\
  SGN \cite{zhang2020semantics} & CVPR20 & 89.0 & 94.5 \\
  4s-Shift-GCN \cite{cheng2020skeleton} & CVPR20 & 90.7 & 96.5 \\
  MS-G3D \cite{liu2020disentangling} & CVPR20 & 91.5 & 96.2 \\
  DC-GCN+ADG \cite{cheng2020decoupling} & ECCV20 & 90.8 & {\bf 96.6} \\
  PA-ResGCN-B19 \cite{song2020stronger} & ACMMM20 & 90.9 & 96.0 \\
  Dynamic-GCN \cite{ye2020dynamic} & ACMMM20 & 91.5 & 96.0 \\
  MST-GCN \cite{chen2021multi} & AAAI21 & 91.5 & {\bf 96.6} \\
  \midrule
  EfficientGCN-B0 & -- & 90.2 & 94.9 \\
  EfficientGCN-B2 & -- & 91.4 & 95.7 \\
  EfficientGCN-B4 & -- & {\bf 92.1} & {\bf 96.1} \\
  \bottomrule
  \end{tabular}
\end{table}

STA-LSTM \cite{song2017end} and AGC-LSTM \cite{si2019attention} are also enhanced by attention modules. However, there are obvious differences between EfficientGCN and these two models, \eg, our attention module works jointly on frames and joints, while their models use the attention modules for frames and joints individually. The performance of EfficientGCN-B4 greatly exceeds STA-LSTM over 10\% on the two benchmarks, and outperforms AGC-LSTM by 2.9\% and 1.1\% on X-sub and X-view benchmarks, respectively. Moreover, 2s-AGCN and its improved versions Dynamic-GCN \cite{ye2020dynamic} can also be considered as a variant of globally spatial attention (non-local structure), while they achieve comparable performances to our model. In addition, DC-GCN+ADG \cite{cheng2020decoupling} utilizes attention mechanism to guide its DropGraph module. This model is better than EfficientGCN-B4 on X-view benchmark, but significantly worse on X-sub benchmark.

\begin{table}[t]
  \caption{Comparisons with SOTA methods on NTU 120 dataset in accuracy (\%). The top part consists of several models without GCN techniques, while the other part contains some graph-based models.}
  \label{tab:ntu120}
  \vspace{-0.4cm}
  \centering
  \setlength{\tabcolsep}{4pt}
  \renewcommand{\arraystretch}{1.2}
  \begin{tabular}{cc|cc}
  \toprule
  Model & Conference & X-sub120 & X-set120 \\
  \midrule
  PA-LSTM \cite{shahroudy2016ntu} & CVPR16 & 25.5 & 26.3 \\
  ST-LSTM \cite{liu2016spatio} & ECCV16 & 55.7 & 57.9 \\
  FSNet \cite{liu2019skeleton} & TPAMI19 & 59.9 & 62.4 \\
  \midrule
  ST-GCN \cite{yan2018spatial} & AAAI18 & 70.7$^\star$ & 73.2$^\star$ \\
  SR-TSL \cite{si2018skeleton} & ECCV18 & 74.1$^\star$ & 79.9$^\star$ \\
  RA-GCNv1 \cite{song2019richly} & ICIP19 & 74.4 & 79.4 \\
  RA-GCNv2 \cite{song2020richly} & TCSVT20 & 81.1 & 82.7 \\
  AS-GCN \cite{li2019actional} & CVPR19 & 77.9$^\star$ & 78.5$^\star$ \\
  2s-AGCN \cite{shi2019two} & CVPR19 & 82.5$^\star$ & 84.2$^\star$ \\
  SGN \cite{zhang2020semantics} & CVPR20 & 79.2 & 81.5 \\
  4s-Shift-GCN \cite{cheng2020skeleton} & CVPR20 & 85.9 & 87.6 \\
  MS-G3D \cite{liu2020disentangling} & CVPR20 & 86.9 & 88.4 \\
  DC-GCN+ADG \cite{cheng2020decoupling} & ECCV20 & 86.5 & 88.1 \\
  PA-ResGCN-B19 \cite{song2020stronger} & ACMMM20 & 87.3 & 88.3 \\
  Dynamic-GCN \cite{ye2020dynamic} & ACMMM20 & 87.3 & 88.6 \\
  MST-GCN \cite{chen2021multi} & AAAI21 & 87.5 & 88.8 \\
  \midrule
  EfficientGCN-B0 & -- & 86.6 & 85.0 \\
  EfficientGCN-B2 & -- & 88.0 & 87.8 \\
  EfficientGCN-B4 & -- & {\bf 88.7} & {\bf 88.9} \\
  \bottomrule
  \multicolumn{4}{l}{$^\star$: These results are implemented based on the released codes.}\\
  \end{tabular}
\end{table}

As an increasingly popular technique, Neural Architecture Search (NAS) has been proposed for automatically searching efficient model structures. The NAS-based methods usually explore all potential topological structures from a large search space, while the hyper-parameters of model blocks (\eg, the width and depth of convolutional layers) are hardly considered. Different from the NAS-based methods, the compound scaling strategy mainly aims to optimize the hyper-parameters of convolutional layers in each model block with fixed topological structures (see Fig.~\ref{fig:model} and Fig.~\ref{fig:layers}), which is easier to implement and cheaper to deploy in real tasks. There have been some studies introducing differentiable NAS into skeleton-based action recognition, \eg, the NAS-GCN \cite{peng2020learning} which has achieved promising results in skeleton action recognition, but is still much more complex than our method, as shown in Tab.~\ref{tab:ntu60}. In the future, it is worthy to study the combination of exploring both topology structures and hyper-parameters for better action recognition models.

It should also be noticed that there are four SOTA models (MS-G3D \cite{liu2020disentangling}, DC-GCN+ADG \cite{cheng2020decoupling}, 4s-Shift-GCN \cite{cheng2020skeleton}, and MST-GCN \cite{chen2021multi}) producing slightly higher accuracies than ours on the X-view benchmark. The inferior performance of the EfficientGCN in cross-view action recognition can be explained from two aspects: 1) in contrast to the 3-stream fusion in the EfficientGCN, the DC-GCN+ADG, the 4s-Shift-GCN and the MST-GCN all adopt 4-stream (joint, bone, motion, and bone motion) fusion strategy, where the additional bone motion information may enhance the robustness of cross-view skeleton features; 2) both the MS-G3D and the MST-GCN introduce multi-scale graph convolution to aggregate more context information in skeleton feature extraction. Though superior performance can be achieved on the X-view benchmark by these models, both the 4-stream fusion strategy and the multi-scale graph convolution definitely increase the model complexities and computational costs.

These results imply that the proposed EfficientGCN is a strong baseline with competitive performance compared to SOTA methods. We consider that this is caused by the superior capability of the compound scaling strategy to balance the model accuracy and complexity, hereby a wider and deeper model can be constructed and easily trained to achieve better performance. Moreover, the proposed ST-JointAtt module contributes to the model accuracy, which makes the model prone to discover the most informative joints.

\subsubsection{NTU RGB+D 120 Dataset}
\label{sssec:ntu120}

As a newly released dataset, there are fewer papers reporting results on the NTU 120 dataset. For comprehensive comparisons, four popular models, \ie, ST-GCN \cite{yan2018spatial}, SR-TSL \cite{si2018skeleton}, AS-GCN \cite{li2019actional} and 2s-AGCN \cite{shi2019two}, are implemented by ourselves, based on their released codes. Tab.~\ref{tab:ntu120} presents the experimental results, from which we can find the proposed EfficientGCN achieves the highest performance, compared to other models. For example, EfficientGCN-B4 outperforms the current SOTA method, MST-GCN \cite{ye2020dynamic}, by 1.2\% on X-sub120 benchmark. Similar to the results on NTU 60 dataset, the performance of EfficientGCN is mainly attributed to the contribution of the compound scaling strategy.

\subsubsection{Model Complexity}
\label{sssec:complexity}

\begin{table}[t]
  \caption{Comparisons with SOTA methods on X-sub benchmark in accuracy (\%), FLOPs ($\times10^9$) and parameter number ($\times10^6$). The models in three parts are compared with EfficientGCN-B0, B2, B4, respectively.}
  \label{tab:complexity}
  \vspace{-0.4cm}
  \centering
  \setlength{\tabcolsep}{4pt}
  \renewcommand{\arraystretch}{1.2}
  \begin{tabular}{cc|cc|cc}
  \toprule
  Model & Acc. & FLOPs & Ratio & \# Param. & Ratio \\
  \midrule
  {\bf EfficientGCN-B0} & {\bf 90.2} & {\bf 2.73} & {\bf 1$\times$} & {\bf 0.29} & {\bf 1$\times$} \\
  ST-GCN \cite{yan2018spatial} & 81.5 & 16.32$^\star$ & 5.98$\times$ & 3.10$^\star$ & 10.69$\times$ \\
  SR-TSL \cite{si2018skeleton} & 84.8 & 4.20$^\star$ & 1.54$\times$ & 19.07$^\star$ & 65.76$\times$ \\
  RA-GCNv1 \cite{song2019richly} & 85.9 & 32.80$^\star$ & 12.01$\times$ & 6.21$^\star$ & 21.41$\times$ \\
  RA-GCNv2 \cite{song2020richly} & 87.3 & 32.80$^\star$ & 12.01$\times$ & 6.21$^\star$ & 21.41$\times$ \\
  AS-GCN \cite{li2019actional} & 86.8 & 26.76$^\star$ & 9.80$\times$ & 9.50$^\star$ & 32.76$\times$ \\
  2s-AGCN \cite{shi2019two} & 88.5 & 37.32$^\star$ & 13.67$\times$ & 6.94$^\star$ & 23.93$\times$ \\
  SGN \cite{zhang2020semantics} & 89.0 & -- & -- & 0.69 & 2.37$\times$ \\
  AGC-LSTM \cite{si2019attention} & 89.2 & -- & -- & 22.89$^\dagger$ & 78.93$\times$ \\
  DGNN \cite{shi2019skeleton} & 89.9 & -- & -- & 26.24$^\dagger$ & 90.48$\times$ \\
  NAS-GCN \cite{peng2020learning} & 89.4 & -- & -- & 6.57$^\dagger$ & 22.66$\times$ \\
  PL-GCN \cite{huang2020part} & 89.2 & -- & -- & 20.70$^\dagger$ & 71.38$\times$ \\
  \midrule
  {\bf EfficientGCN-B2} & {\bf 91.4} & {\bf 4.05} & {\bf 1$\times$} & {\bf 0.51} & {\bf 1$\times$} \\
  4s-Shift-GCN \cite{cheng2020skeleton} & 90.7 & 10.0 & 2.47$\times$ & 2.76$^\star$ & 5.41$\times$ \\
  DC-GCN+ADG \cite{cheng2020decoupling} & 90.8 & 25.72$^\star$ & 6.35$\times$ & 4.96$^\star$ & 9.73$\times$ \\
  PA-ResGCN-B19 \cite{song2020stronger} & 90.9 & 18.52$^\star$ & 4.57$\times$ & 3.64 & 7.14$\times$ \\
  \midrule
  {\bf EfficientGCN-B4} & {\bf 92.1} & {\bf 8.36} & {\bf 1$\times$} & {\bf 1.10} & {\bf 1$\times$} \\
  MS-G3D \cite{liu2020disentangling} & 91.5 & 48.88$^\star$ & 5.85$\times$ & 6.40 & 5.82$\times$ \\
  Dynamic-GCN \cite{ye2020dynamic} & 91.5 & -- & -- & 14.40$^\dagger$ & 13.09$\times$ \\
  MST-GCN \cite{chen2021multi} & 91.5 & -- & -- & 12.00 & 10.91$\times$ \\
  \bottomrule
  \multicolumn{6}{l}{$^\star$: These results are implemented based on the released codes.}\\
  \multicolumn{6}{l}{$^\dagger$: These results are provided by their authors.}\\
  \end{tabular}
\end{table}

In order to verify the efficiency of our model, we compare our EfficientGCN family with other methods in terms of accuracy and model complexity (FLOPs and number of parameters) on X-sub benchmark of NTU 60 dataset. The experimental results are presented in Tab.~\ref{tab:complexity}. This table is divided into three parts, where the models are grouped into three parts with different accuracies. The ratios following FLOPs and parameter number denote the ratio between the model and its corresponding EfficientGCN. Due to the lack of reported FLOPs and parameter numbers for most models, we obtain the results by their released codes or directly asking their authors for helps. Note that the FLOPs of SGN \cite{zhang2020semantics} and Dynamic-GCN \cite{ye2020dynamic} are reported by their authors but not presented in this table, because these two models contain well-designed data transformation modules, which resize the original skeleton sequence to a very short sequence (\eg, 20 frames), instead of performing on the whole 300 frames. Thus, we ignore the FLOPs of these two models and only give the numbers of their parameters for fair comparisons.

In the top part of Tab.~\ref{tab:complexity}, it is observed that there is a large gap between the efficiencies of EfficientGCN-B0 and previous models. Compared to the first GCN baseline for skeleton-based action recognition, \ie, ST-GCN \cite{yan2018spatial}, EfficientGCN-B0 outperforms by 8.7\% in accuracy, with a 5.98$\times$ fewer FLOPs and a 10.68$\times$ fewer parameters. DGNN \cite{shi2019skeleton} obtains the same accuracy as EfficientGCN-B0, but its amount of trainable parameters is exceedingly larger, about 90$\times$ than EfficientGCN-B0. SGN \cite{zhang2020semantics} is a lightweight and efficient model for skeleton-based action recognition, which achieves 89.0\% accuracy with only 0.69$\times10^6$ parameters. However, it is still worse than our EfficientGCN-B0 in both model accuracy and model size.

As to the middle part, EfficientGCN-B2 achieves 91.4\% accuracy with 4.05$\times10^9$ FLOPs and 0.51$\times10^6$ parameters, which are about 4.57$\times$ faster and 7.14$\times$ smaller than the preliminary version of this paper (PA-ResGCN-B19). Furthermore, the bottom part shows that EfficientGCN-B4 achieves a SOTA accuracy with a small amount of trainable parameters. Though it has been around 4$\times$ larger than the EfficientGCN-B0, it is still much fewer than the other models with similar performance. These results clearly show that the proposed method brings a remarkable improvement in both model accuracy and complexity, which will benefit to the real applications of skeleton-based action recognition.

\begin{figure}[t]
  \centerline{\includegraphics[width=9cm]{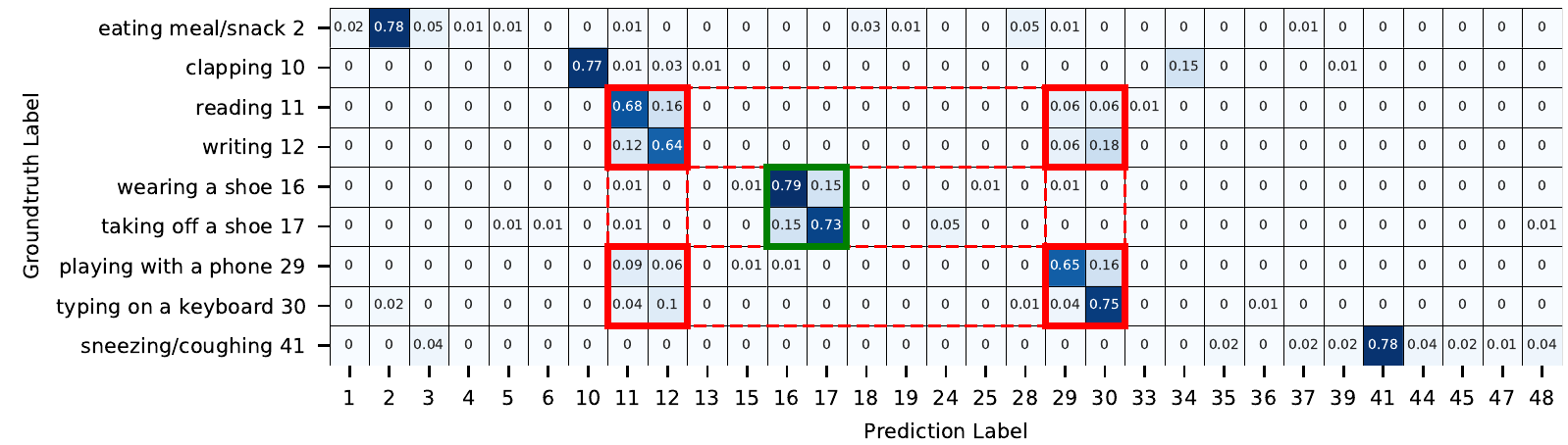}}
  \centerline{\includegraphics[width=9cm]{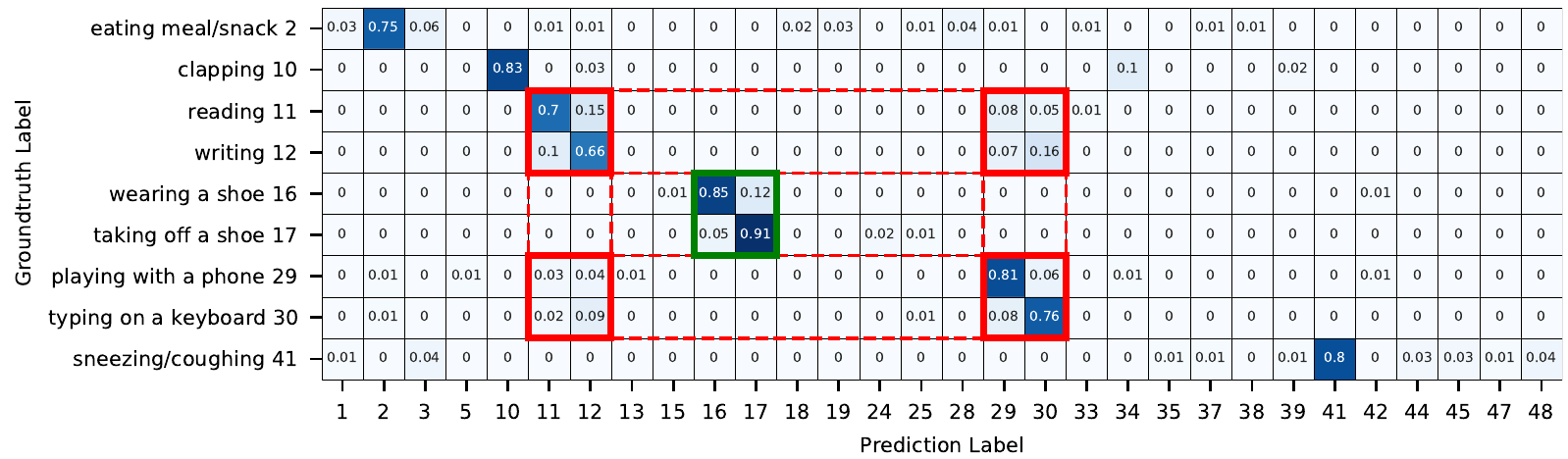}}
  \vspace{-0.4cm}
  \caption{Confusion matrices of EfficientGCN-B0 (top) and EfficientGCN-B4 (bottom) with failure actions (less than 80\% accuracy on X-sub benchmark), where the numbers in coordinate axes represent the indexes of each action category, the {\color{red}red} and {\color{darkgreen}green} rectangles denote two groups of similar actions. \bv}\label{fig:cm}
\end{figure}

\subsection{Discussion and Visualization}
\label{ssec:discuss}

\subsubsection{Confusion Matrices and Failure Cases}
\label{sssec:failure}

Although EfficientGCN receives promising results on the large-scale datasets, there are still some actions difficult to be well recognized. As Fig.~\ref{fig:cm} displays, we draw the confusion matrices of some actions for the proposed EfficientGCN-B0 and EfficientGCN-B4, respectively, where the selected actions are determined according to their insufficient accuracies (less than 80\% on X-sub benchmark). From the top row of Fig.~\ref{fig:cm}, two groups of similar actions should be noticed. The first one is marked by the red rectangles, including {\it reading}, {\it writing}, {\it playing with a phone}, and {\it typing on a keyboard}. All these actions are mainly performed by the slight shaking of two hands, which are extremely similar at spatial configurations and temporal dynamics. The second group, surrounded by green rectangles, consists of two similar actions, \ie, {\it wearing a shoe} and {\it taking off a shoe}. These two actions have similar spatial configurations, but different temporal dynamics. With respect to EfficientGCN-B4 (bottom row of Fig.~\ref{fig:cm}), the recognition accuracies of actions in the second group receive a significant improvement, while the actions in the first group are still hard to be distinguished.

\begin{figure*}[t]
  \vspace{-0.2cm}
  \centerline{\includegraphics[width=18cm]{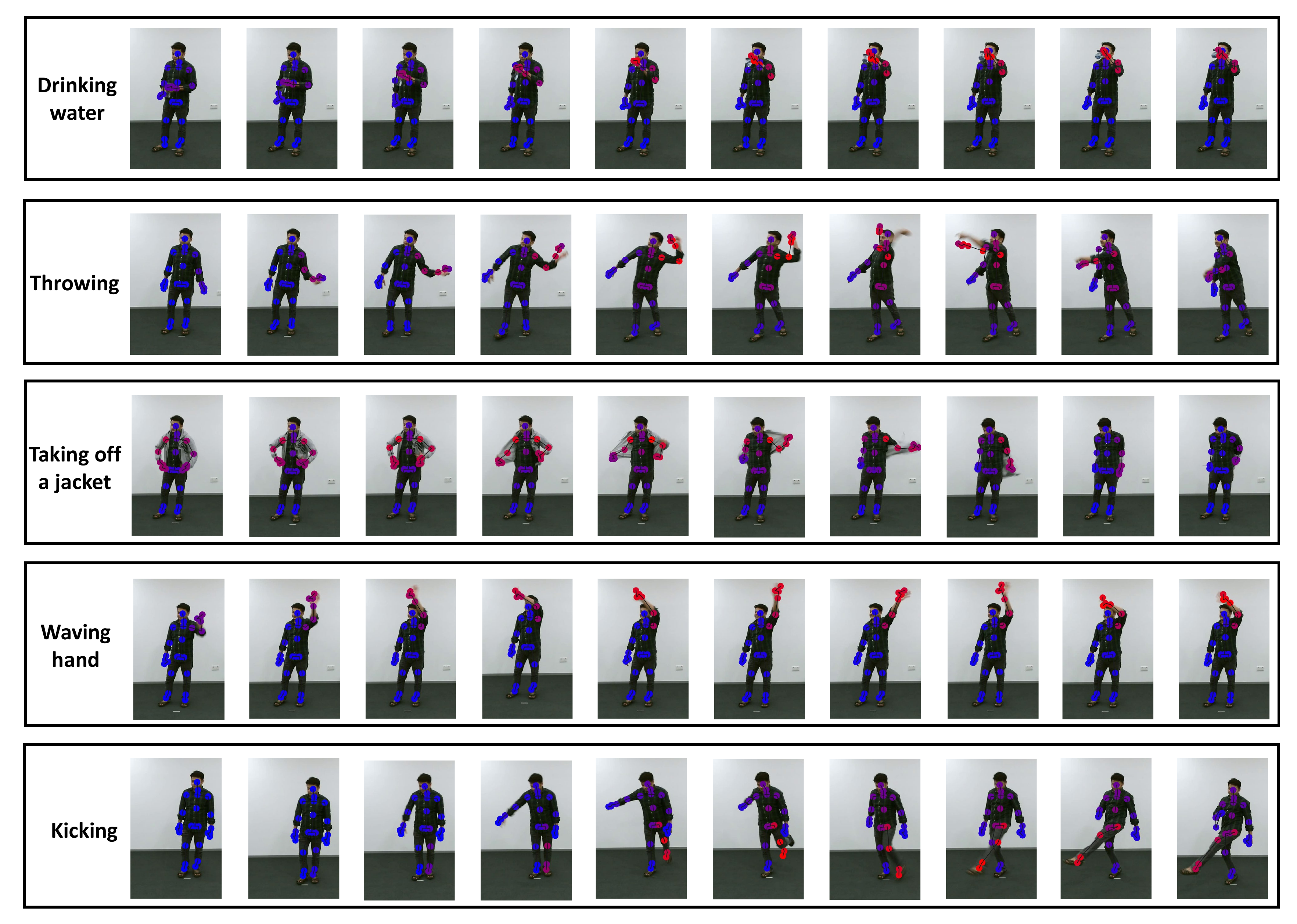}}
  \vspace{-0.4cm}
  \caption{Activated joints in 10 contextual frames of EfficientGCN-B4 for the sample actions, \ie, {\it drinking water}, {\it throwing}, {\it taking off a jacket}, {\it waving hand}, and {\it kicking}. The {\color{red}red} points denote the activated joints, while {\color{blue}blue} points represent non-activated joints. \bv}\label{fig:samples}
  \vspace{-0.2cm}
\end{figure*}

This issue is mainly caused by the lack of joints to represent two hands, thus the information of two hands is generally insufficient. Furthermore, the fringe joints of NTU 60 dataset often contain much noise (\eg, the 4th, 7th, and 8th frames of {\it throwing} in Fig.~\ref{fig:samples}), which makes a huge influence on capturing the discriminative features. However, our approach is not particularly designed for dealing with noises. Therefore, it is still challenging to recognize such subtle actions.

\begin{figure}[t]
  \vspace{-0.2cm}
  \centerline{\includegraphics[width=9cm]{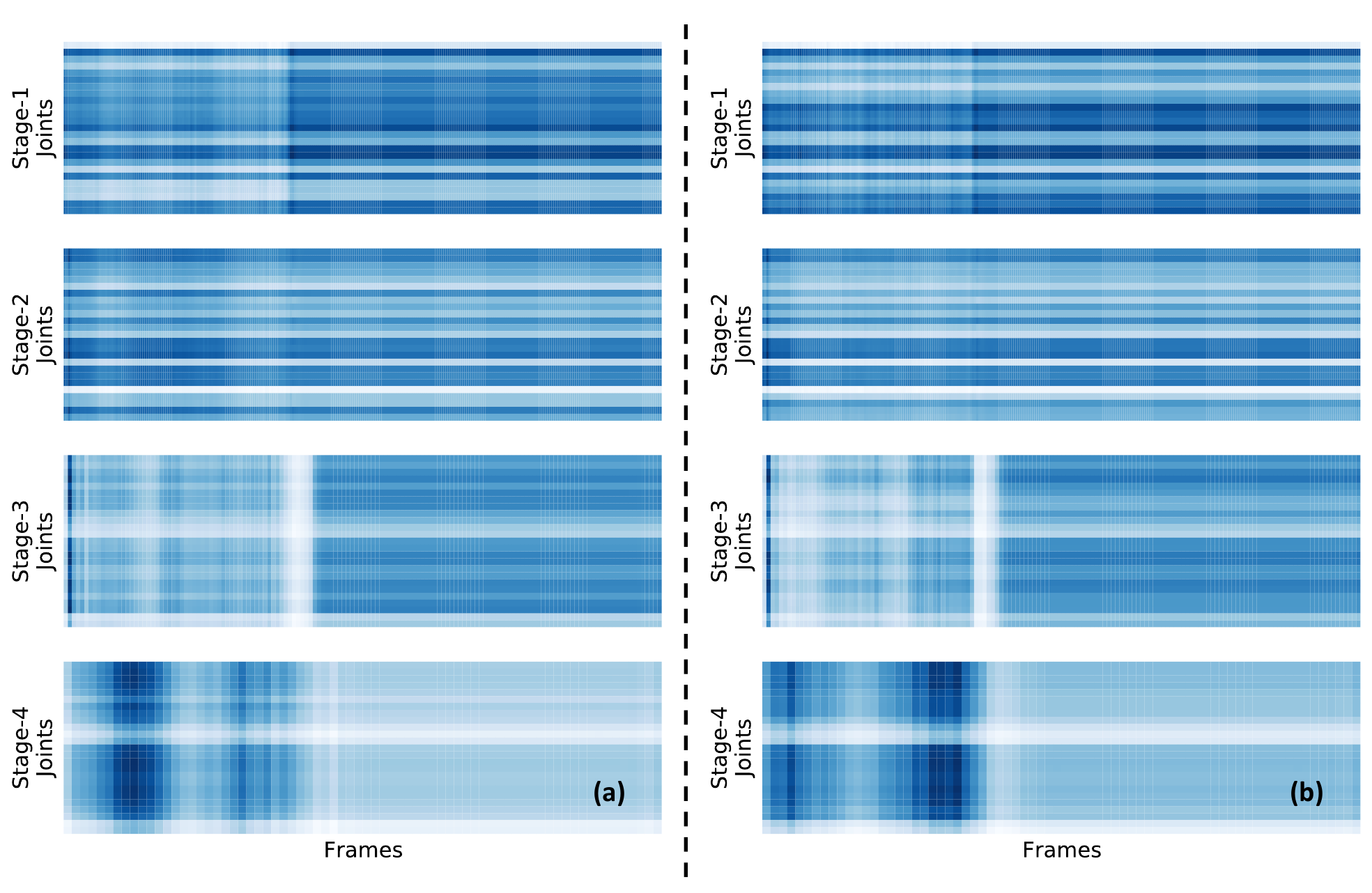}}
  \vspace{-0.4cm}
  \caption{The subfigures (a) and (b) describe the attention maps of two randomly selected samples respectively, where the four parts of each subfigure are calculated by the attention modules in four stages of EfficientGCN-B4 model. A small square with darker color denotes a higher attention weight for the corresponding spatial temporal joint. All these attention maps are performed on X-sub benchmark. \bv}\label{fig:att_maps}
\end{figure}

\subsubsection{Attention Maps}

To illustrate the characteristics of the ST-JointAtt module, we depict the attention maps of two randomly selected samples at four stages of the EfficientGCN-B4 model. As shown in Fig.~\ref{fig:att_maps}, for the two different action samples, the left and right subfigures share similar attention maps at all stages. For each subfigure, the top two feature maps at the first two GCN stages show the stronger selectivity on spatial joints, and the bottom two at high-level GCN stages display additional selectivity on time frames. It should be noticed that the frames in the later of the sequence which are out of the duration of actions are padded by zeros, the attention maps at the last stage successfully distinguish these uninformative frames with small weights. From these figures, it shows that the proposed ST-JointAtt module pays more attention to informative joints at the early convolutional stages, while distinguishes informative frames in the later convolutional stages. Meanwhile, the joint sensitivity is weakened at the later stages, which may be caused by the over-smoothing problem in GCN as the adjacent skeleton joints tend to become indistinguishable in deeper layers.

\subsubsection{Class Activation Maps}
\label{sssec:cam}

To show how our model works, the activation maps of some skeleton sequences are calculated by class activation map \cite{learning2016zhou}, as presented in Fig.~\ref{fig:samples}, in which the activated joints in several sampled frames are displayed. From this figure, we can find that the EfficientGCN-B4 model successfully concentrates on the most informative joints, \ie, left arm for {\it drinking water}, {\it throwing} and {\it waving hand}, upper body for {\it taking off a jacket}, and left leg for {\it kicking}. This implies that the proposed ST-JointAtt module works well. Besides, compared with the preliminary PartAtt module proposed in \cite{song2020stronger}, this new attention module results in more reasonable attention weights in temporal dimension, by which only informative moving joints in certain frames are captured (\eg, all joints in the first two frames of {\it kicking} are not activated).

\subsection{Generalization of EfficientGCN}
\label{ssec:generalization}

From the above experimental results, the EfficientGCN shows an outstanding performance for the skeleton-based action recognition task. To further validate the generalization ability of the EfficientGCN, we apply the EfficientGCN-B4 to other skeleton-based tasks such as person re-identification (ReID) \cite{zhao2017spindle,rao2021self}.

Following the procedure in \cite{rao2021self}, four open benchmarks on skeleton-based person ReID are employed, including BIWI \cite{munaro2014one}, IAS-A/IAS-B \cite{munaro2014feature}, and KGBD \cite{andersson2015person}. The same training/testing splits for each dataset are adopted as mentioned in \cite{rao2021self}, and the number of input frame is set to 80 for KGBD and 10 for others in this paper. Note that the skeleton sequences with more than 80 (or 10) frames will be split to several samples with the same person ID. The experimental results on the four benchmarks are displayed in Tab.~\ref{tab:generalization}. It can be seen that the EfficientGCN-B4 model significantly outperforms the other SOTA models on three benchmarks, indicating that our model has an excellent performance on skeleton-based person ReID task, further showing its great potential to other skeleton-based motion analysis tasks.

\begin{table}[t]
  \caption{Comparisons with other person ReID methods in rank-1 accuracy (\%).}
  \label{tab:generalization}
  \centering
  \setlength{\tabcolsep}{4pt}
  \renewcommand{\arraystretch}{1.2}
  \vspace{-0.2cm}
  \begin{tabular}{c|cccc}
  \toprule
  Method & BIWI & IAS-A & IAS-B & KGBD \\
  \midrule
  $D^{13}$+KNN \cite{munaro2014one} & 39.3 & 33.8 & 40.5 & 46.9 \\
  $D^{16}$+Adaboost \cite{pala2019enhanced} & 41.8 & 27.4 & 39.2 & 69.9 \\
  Single-layer LSTM \cite{haque2016recurrent} & 15.8 & 20.0 & 19.1 & 39.8 \\
  Multi-layer LSTM \cite{zheng2019relational} & 36.1 & 34.4 & 30.9 & 46.2 \\
  PoseGait \cite{liao2020model} & 33.3 & 41.4 & 37.1 & 90.6 \\
  Locality-Awareness-SGE \cite{rao2021self} & 63.3 & 60.1 & {\bf 62.5} & 90.6 \\
  \midrule
  EfficientGCN-B4 & {\bf 64.5} & {\bf 67.3} & {\bf 62.4} & {\bf 96.1} \\
  \bottomrule
  \end{tabular}
\end{table}

\section{Conclusion}
\label{sec:conclusion}

In this paper, we have constructed a family of efficient but strong baselines based on a set of techniques for boosting model efficiencies. Different from other multi-stream models, the proposed EfficientGCN fuses three input branches at early stage, obviously eliminating the redundant parameters. In order to further reduce the model complexity, four TC layers are designed according to the bottleneck structure and separable convolution, which significantly saves the computational cost. Moreover, a compound scaling strategy is utilized to uniformly scale the model width and depth, further reducing the model complexity. On two large-scale datasets, NTU RGB+D 60 \& 120, the proposed EfficientGCN-B4 achieves the SOTA performance, while its FLOPs and number of parameters are obviously fewer than other models. Thus, the new baselines will have a huge potential for developing more complicated models. In the future, we will extend the proposed baseline with the object appearance, which is likely responsible for the recognition of some extremely similar actions.

\section*{Acknowledgments}

This work is sponsored by the National Key R\&D Program of China under Grant 2016YFB1001002, the National Natural Science Foundation of China under Grant 61525306, Grant 61633021, and Grant 61721004, the Shandong Provincial Key Research and Development Program (Major Scientific and Technological Innovation Project) under Grant 2019JZZY010119, and CAS-AIR.

{\small
  \bibliographystyle{IEEEtran}
  \bibliography{conferences_abrv,references}
}

\begin{IEEEbiography}[{\includegraphics[width=1in,height=1.25in,clip,keepaspectratio]{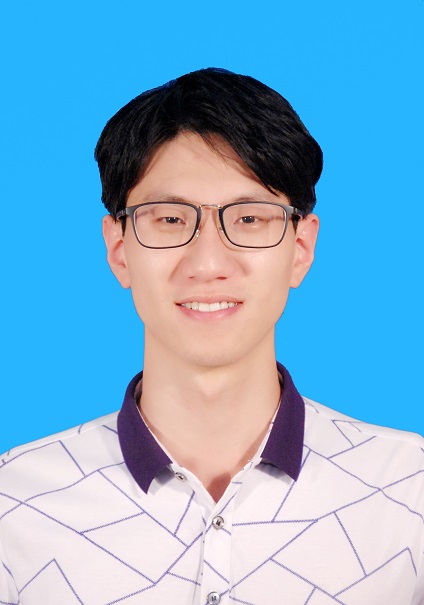}}]{Yi-Fan Song}
received the M.Eng. degree from Zhengzhou University, Zhengzhou, China, in 2018. Currently, He is a Ph.D. candidate of the School of Artificial Intelligence, University of Chinese Academy and Sciences (UCAS). His research interests include computer vision, action recognition, action prediction, and neural architecture search.
\end{IEEEbiography}

\begin{IEEEbiography}[{\includegraphics[width=1in,height=1.25in,clip,keepaspectratio]{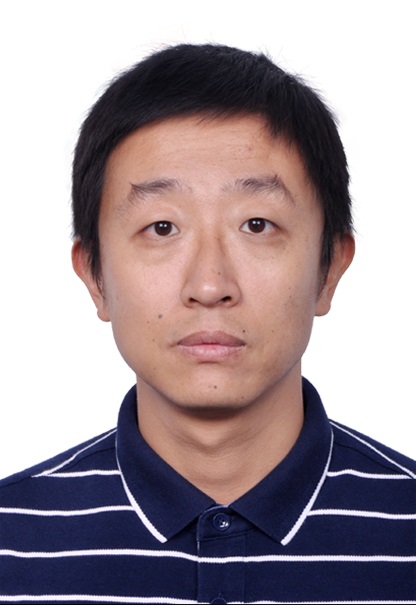}}]{Zhang Zhang}
received the B.S. degree in computer science and technology from Hebei University of Technology, Tianjin, China, in 2002, and the Ph.D. degree in pattern recognition and intelligent systems from the National Laboratory of Pattern Recognition, Institute of Automation, Chinese Academy of Sciences, Beijing, China in 2009. Currently, he is an associate professor at the National Laboratory of Pattern Recognition, Institute of Automation, Chinese Academy of Sciences (CASIA). His research interests include activity recognition, video surveillance, and time series analysis. He has published 20s research papers on computer vision and pattern recognition, including IEEE TPAMI, CVPR, and ECCV etc.
\end{IEEEbiography}

\begin{IEEEbiography}[{\includegraphics[width=1in,height=1.25in,clip,keepaspectratio]{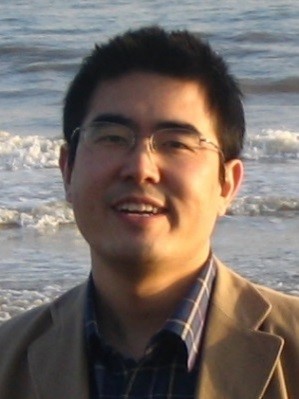}}]{Caifeng Shan}
received the B.Eng. degree from the University of Science and Technology of China (USTC), the M.Eng. degree from the Institute of Automation, Chinese Academy of Sciences, and the Ph.D. degree in computer vision from Queen Mary, University of London. His research interests include computer vision, pattern recognition, image and video analysis, machine learning, bio-medical imaging, and related applications. He has authored more than 100 papers and 60 patent applications. He has served as Associate Editor or Guest Editor for many scientific journals including IEEE Transactions on Circuits and Systems for Video Technology and IEEE Journal of Biomedical and Health Informatics. He is a Senior Member of IEEE.
\end{IEEEbiography}

\begin{IEEEbiography}[{\includegraphics[width=1in,height=1.25in,clip,keepaspectratio]{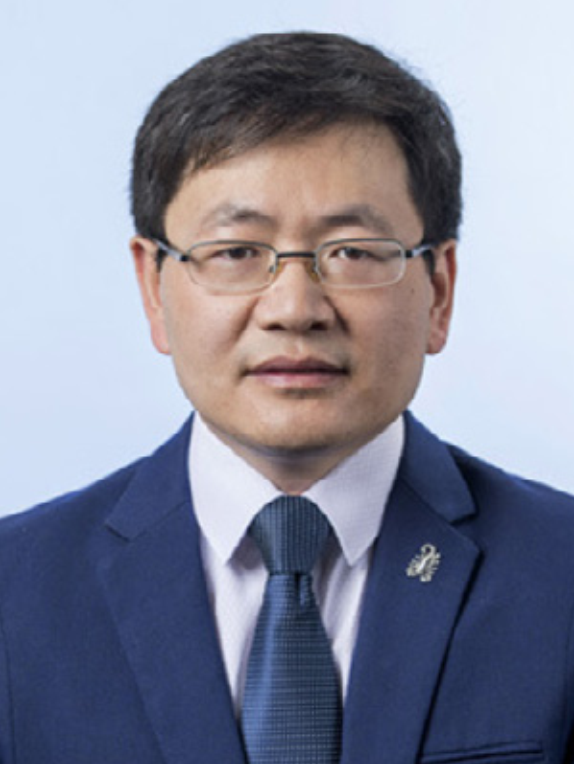}}]{Liang Wang}
received both the B.Eng. and M.Eng. degrees from Anhui University in 1997 and 2000, respectively, and the Ph.D. degree from the Institute of Automation, Chinese Academy of Sciences (CASIA) in 2004. From 2004 to 2010, he was a research assistant at Imperial College London, United Kingdom, and Monash University, Australia, a research fellow at the University of Melbourne, Australia, and a lecturer at the University of Bath, United Kingdom, respectively. Currently, he is a full professor of the Hundred Talents Program at the National Lab of Pattern Recognition, CASIA. His major research interests include machine learning, pattern recognition, and computer vision. He has widely published in highly ranked international journals such as IEEE Transactions on Pattern Analysis and Machine Intelligence and IEEE Transactions on Image Processing, and leading international conferences such as CVPR, ICCV, and ICDM. He is an IEEE Fellow, and an IAPR Fellow.
\end{IEEEbiography}

\newpage
\appendices
\setcounter{page}{1}
\setcounter{equation}{0}
\renewcommand\theequation{S.\arabic{equation}}
\setcounter{table}{0}
\renewcommand\thetable{S.\arabic{table}}
\setcounter{figure}{0}
\renewcommand\thefigure{S.\arabic{figure}}

\section{Network Architecture}
\label{asec:architectures}

Here we present the details to calculate the scaled width and depth. Firstly, the initial channels and TC layers of four blocks are set to $\{48,24,64,128\}$ and $\{0.5,0.5,1,1\}$, respectively. Then, the rounding functions are given as:
\begin{equation}
  C_\phi=\epsilon(C_0/16\times\alpha^\phi)*16
\end{equation}
\begin{equation}
  L_\phi=\epsilon(L_0\times\beta^\phi)
\end{equation}
where $C_\phi$ and $L_\phi$ denote the numbers of scaled channels and TC layers with scaling coefficient $\phi$, $C_0$ and $L_0$ denote the initial channels and TC layers, $\alpha$ and $\beta$ are defined in Sec.~4.3, and $\epsilon(\cdot)$ represents the step function formulated as:
\begin{equation}
  \epsilon(x)=\left\{\begin{matrix}
    \lceil x \rceil, & if \quad x-\lfloor x \rfloor>0.5 \\
    \lfloor x \rfloor, & if \quad x-\lfloor x \rfloor\leq0.5
  \end{matrix}\right.
\end{equation}
where $\lceil \cdot \rceil$ and $\lfloor \cdot \rfloor$ mean up and down rounding functions, respectively.

Finally, the network architectures of models with scaling coefficients $\{2,4\}$, \ie, EfficientGCN-B2 and EfficientGCN-B4, can be calculated by initial channels, initial TC layers, and rounding functions. The architectures of these three baselines are shown in Tab.~\ref{tab:architecture0}, Tab.~\ref{tab:architecture2} and Tab.~\ref{tab:architecture4}.

\begin{table}[ht]
  \vspace{-0.2cm}
  \caption{The architecture of EfficientGCN-B0 network. Each row describes a block with the following settings and output shape, where $\times3$ represents three input branches, $/2$ denotes a stride of 2, $Q$ is the number of action classes.}
  \label{tab:architecture0}
  \vspace{-0.4cm}
  \centering
  \setlength{\tabcolsep}{4pt}
  \renewcommand{\arraystretch}{1.2}
  \begin{tabular}{ccccc}
  \toprule
  Stage & Block & Depth & Channels & Shape \\
  \midrule
  -- & BN$\times3$ & -- & $(6,6)\times3$ & $(6\times T_{in}\times V_{in})\times3$ \\
  -- & $^\star$BasicBlock$\times3$ & 1 & $(6,64)\times3$ & $(64\times T_{in}\times V_{in})\times3$ \\
  \hline
  1 & SGBlock$\times3$ & 0$^\dagger$ & $(64,48)\times3$ & $(48\times T_{in}\times V_{in})\times3$ \\
  2 & SGBlock$\times3$ & 0$^\dagger$ & $(48,16)\times3$ & $(16\times T_{in}\times V_{in})\times3$ \\
  \hline
  -- & Concat & -- & $16\times3, 48$ & $48\times T_{in}\times V_{in}$ \\
  \hline
  3 & SGBlock & 1 & $48,64,/2$ & $64\times T_{in}/2\times V_{in}$ \\
  4 & SGBlock & 1 & $64,128,/2$ & $128\times T_{in}/4\times V_{in}$ \\
  \hline
  -- & GAP & -- & $128,128$ & $128$ \\
  \hline
  -- & FC & -- & $128,Q$ & $Q$ \\
  \bottomrule
  \multicolumn{5}{l}{$^\star$: This BasicBlock is fixed and without attentions for stable training.}\\
  \multicolumn{5}{l}{$^\dagger$: Actually, the initial depths of the two blocks in input branches are}\\
  \multicolumn{5}{l}{both 0.5. The depth of 0 is obtained after rounding.}\\
  \end{tabular}
  \vspace{-0.2cm}
\end{table}

\begin{table}[ht]
  \vspace{-0.4cm}
  \caption{The architecture of EfficientGCN-B2 network. Each row describes a block with the following settings and output shape, where $\times3$ represents three input branches, $/2$ denotes a stride of 2, $Q$ is the number of action classes.}
  \label{tab:architecture2}
  \vspace{-0.4cm}
  \centering
  \setlength{\tabcolsep}{4pt}
  \renewcommand{\arraystretch}{1.2}
  \begin{tabular}{ccccc}
  \toprule
  Stage & Block & Depth & Channels & Shape \\
  \midrule
  -- & BN$\times3$ & -- & $(6,6)\times3$ & $(6\times T_{in}\times V_{in})\times3$ \\
  -- & $\star$BasicBlock$\times3$ & 1 & $(6,64)\times3$ & $(64\times T_{in}\times V_{in})\times3$ \\
  \hline
  1 & SGBlock$\times3$ & 1 & $(64,64)\times3$ & $(64\times T_{in}\times V_{in})\times3$ \\
  2 & SGBlock$\times3$ & 1 & $(64,32)\times3$ & $(32\times T_{in}\times V_{in})\times3$ \\
  \hline
  -- & Concat & -- & $32\times3, 96$ & $96\times T_{in}\times V_{in}$ \\
  \hline
  3 & SGBlock & 2 & $96,96,/2$ & $96\times T_{in}/2\times V_{in}$ \\
  4 & SGBlock & 2 & $96,192,/2$ & $192\times T_{in}/4\times V_{in}$ \\
  \hline
  -- & GAP & -- & $192,192$ & $192$ \\
  \hline
  -- & FC & -- & $192,Q$ & $Q$ \\
  \bottomrule
  \multicolumn{5}{l}{$^\star$: This BasicBlock is fixed and without attentions for stable training.}\\
  \end{tabular}
\end{table}

\begin{table}[ht]
  \caption{The architecture of EfficientGCN-B4 network. Each row describes a block with the following settings and output shape, where $\times3$ represents three input branches, $/2$ denotes a stride of 2, $Q$ is the number of action classes.}
  \label{tab:architecture4}
  \vspace{-0.4cm}
  \centering
  \setlength{\tabcolsep}{4pt}
  \renewcommand{\arraystretch}{1.2}
  \begin{tabular}{ccccc}
  \toprule
  Stage & Block & Depth & Channels & Shape \\
  \midrule
  -- & BN$\times3$ & -- & $(6,6)\times3$ & $(6\times T_{in}\times V_{in})\times3$ \\
  -- & $\star$BasicBlock$\times3$ & 1 & $(6,64)\times3$ & $(64\times T_{in}\times V_{in})\times3$ \\
  \hline
  1 & SGBlock$\times3$ & 2 & $(64,96)\times3$ & $(96\times T_{in}\times V_{in})\times3$ \\
  2 & SGBlock$\times3$ & 2 & $(96,48)\times3$ & $(48\times T_{in}\times V_{in})\times3$ \\
  \hline
  -- & Concat & -- & $48\times3, 144$ & $144\times T_{in}\times V_{in}$ \\
  \hline
  3 & SGBlock & 3 & $144,128,/2$ & $128\times T_{in}/2\times V_{in}$ \\
  4 & SGBlock & 3 & $128,272,/2$ & $272\times T_{in}/4\times V_{in}$ \\
  \hline
  -- & GAP & -- & $272,272$ & $272$ \\
  \hline
  -- & FC & -- & $272,Q$ & $Q$ \\
  \bottomrule
  \multicolumn{5}{l}{$^\star$: This BasicBlock is fixed and without attentions for stable training.}\\
  \end{tabular}
\end{table}

\section{Grid Search for Receptive Field}
\label{asec:receptive}

There are two hyper-parameters mentioned in Sec.~\ref{ssec:graphconv} and \ref{ssec:block}, \ie, $D$ for the maximum spatial graph distance and $L$ for the temporal window size. They directly determine the receptive field of the base convolutional operation, thus have implication on the model performance. The effects of these two hyper-parameters are discussed in this section, and the experimental results are shown in Tab.~\ref{tab:receptive_acc} and Tab.~\ref{tab:receptive_complexity}. For the maximum graph distance $D$, we can find that there is an obvious decline when $D<2$. However, $D>3$ is not better than $D=2$ or $D=3$ because an oversized receptive field will make the skeleton graph over-smoothing. Similarly, the temporal window size $L$ is also required to choose a suitable value by balancing the model accuracy and complexity. Thus, according to these two tables, we set $D=2$ and $L=5$ ({\bf Bold} in tables) by choosing a high model accuracy and a low model complexity simultaneously. Note that the settings of $D=2,L=11$ and $D=3,L=9$ are slightly more accurate than $D=2,L=5$, but their model complexities are considerably higher than the selected setting.

\begin{table}[ht]
  \caption{Comparisons with different receptive fields on X-sub benchmark in accuracy (\%).}
  \label{tab:receptive_acc}
  \vspace{-0.4cm}
  \centering
  \setlength{\tabcolsep}{4pt}
  \renewcommand{\arraystretch}{1.2}
  \begin{tabular}{c|ccccc}
  \toprule
  Acc. & $D=1$ & $D=2$ & $D=3$ & $D=4$ & $D=5$ \\
  \midrule
  $L=3$ & 87.7 & 88.9 & 89.2 & 88.8 & 88.5 \\
  $L=5$ & 88.1 & {\bf 89.9} & 89.9 & 89.7 & 89.5 \\
  $L=7$ & 88.6 & 89.8 & 89.9 & 89.5 & 89.7 \\
  $L=9$ & 88.8 & 89.9 & 90.0 & 89.8 & 89.9 \\
  $L=11$ & 88.7 & 90.0 & 89.9 & 89.7 & 89.7 \\
  \bottomrule
  \end{tabular}
\end{table}

\begin{table}[ht]
  \vspace{-0.4cm}
  \caption{Comparisons with different receptive fields on X-sub benchmark in FLOPs ($\times10^9$) and parameter numbers ($\times10^6$).}
  \label{tab:receptive_complexity}
  \vspace{-0.4cm}
  \centering
  \setlength{\tabcolsep}{4pt}
  \renewcommand{\arraystretch}{1.2}
  \begin{tabular}{c|ccccc}
  \toprule
  F. / P.& $D=1$ & $D=2$ & $D=3$ & $D=4$ & $D=5$ \\
  \midrule
  $L=3$ & 2.35/0.27 & 2.72/0.30 & 3.09/0.33 & 3.46/0.36 & 3.83/0.39 \\
  $L=5$ & 2.70/0.29 & {\bf 3.08/0.32} & 3.45/0.35 & 3.82/0.38 & 4.19/0.41 \\
  $L=7$ & 3.06/0.32 & 3.43/0.35 & 3.80/0.38 & 4.18/0.41 & 4.55/0.44 \\
  $L=9$ & 3.42/0.35 & 3.79/0.38 & 4.16/0.41 & 4.53/0.44 & 4.90/0.47 \\
  $L=11$ & 3.78/0.37 & 4.15/0.40 & 4.52/0.43 & 4.89/0.46 & 5.26/0.49 \\
  \bottomrule
  \end{tabular}
\end{table}

\end{document}